%% file: decisiontracebench.tex
\documentclass[11pt]{article}
\usepackage[preprint]{acl}
\usepackage{times,latexsym}
\usepackage[T1]{fontenc}
\usepackage[utf8]{inputenc}
\usepackage{microtype}
\usepackage{amsmath,amssymb,booktabs,tabularx,array,fvextra,placeins,xurl,graphicx}
\usepackage{float,needspace}
\newcommand{\bench}{DecisionTraceBench}
\title{DecisionTraceBench: Tracking Decision-Critical Information in Multi-Party Spoken Conversations}
\author{Wenchao Du$^{1,2}$ \quad Fangxin Shang$^{1}$ \quad Long Wu$^{1}$ \quad Runze Cui$^{1}$ \quad Qing Yang$^{1}$ \quad Yehui Yang$^{1,*}$ \\
  \normalfont $^{1}$AI Lab, Qifu Technology, Beijing, China \\
  \normalfont $^{2}$University of Chinese Academy of Sciences \\
  \normalfont\small $^{*}$Corresponding author: \href{mailto:yangyehuisw@126.com}{\texttt{yangyehuisw@126.com}}}
\hypersetup{
  pdftitle={DecisionTraceBench: Tracking Decision-Critical Information in Multi-Party Spoken Conversations},
  pdfauthor={Wenchao Du, Fangxin Shang, Long Wu, Runze Cui, Qing Yang, Yehui Yang}
}
\begin{document}

\raggedbottom
\maketitle

\begin{abstract}
In high-stakes domains such as finance, critical information evolves through multi-party interaction. Recovering these changes and their sources is fundamental to reliable speech understanding. Existing speech benchmarks focus on transcription, local understanding, and turn-level state prediction, leaving joint decision evolution across speakers insufficiently evaluated. We introduce DecisionTraceBench, a speech state-tracking benchmark for Decision-Critical Information State Tracking (DCIST). It jointly evaluates decision event sequences, speaker attribution, scope, and temporal evidence, using trajectory-level F1 to measure complete event-chain recovery. Our dataset comprises Chinese financial dialogues involving two to four speakers, with 100.39 hours of synthetic audio and 8.61 hours of human-performed recordings. On a fixed cohort of human-performed recordings, five cascaded systems and two direct-audio models achieve DCIST scores of 5.73\% to 21.35\%. Ablations with speaker-labeled reference scripts and no temporal localization requirements yield a maximum score of only 38.72\%. Audio perturbation experiments further show that overall transcription error rates do not adequately reflect complete trajectory recovery. These findings reveal limitations in jointly understanding information provenance and state evolution, providing a benchmark for evaluating decision-process understanding and decision traceability in high-stakes speech applications.
\end{abstract}

\begin{figure*}[t]
\centering
\includegraphics[width=\textwidth]{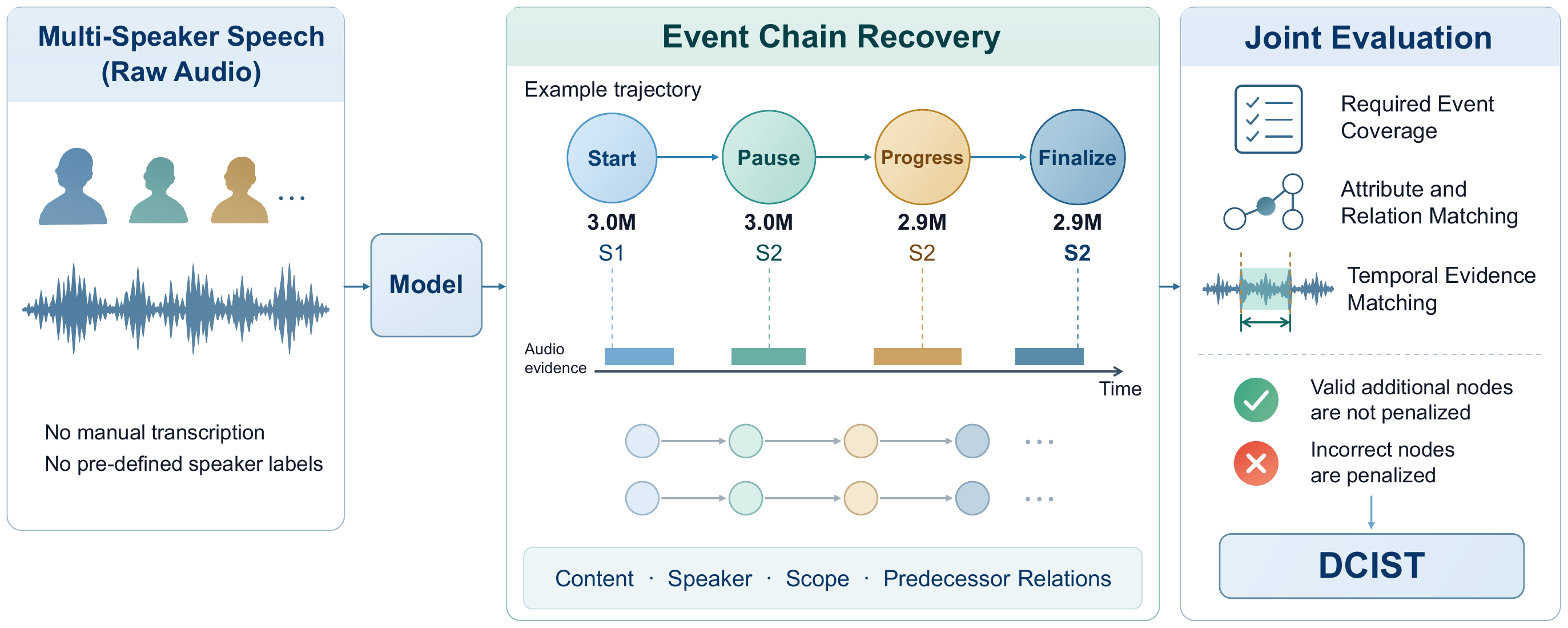}
\caption{Full-trajectory speech state tracking. Systems recover an event chain from audio without manual transcripts or gold speaker labels. Evaluation jointly checks required-event coverage, attributes and relations, and temporal evidence. Start, Pause, Progress, and Finalize correspond to \texttt{start}, \texttt{hold}, \texttt{advance}, and \texttt{settle}.}\label{fig:overview}
\end{figure*}

\section{Introduction}
In financial decision-making, a conversational record must establish not only what was said, but how a proposal was revised, suspended, resumed, and resolved, and who was responsible for each effective change. Identical terminal values can arise through different processes. The same proposition can also have different force when expressed by a participant making a request, an analyst verifying evidence, or an authorized decision-maker. Recovering this history is essential to a traceable record.

Existing benchmarks evaluate broad audio understanding, speaker-attributed reasoning, and turn-level dialogue state tracking \citep{r1,r6,r7,r8}. Text-based resources also model successive state changes \citep{propara,openpi}. Yet local correctness does not establish trajectory correctness: a model may recover the terminal amount while omitting a suspension and resumption, or recognize every amount while attributing another participant's suggestion to an authorized decision-maker. Full-trajectory recovery requires jointly identifying effective events, their relations, their sources, and their evidence.

We introduce \bench, a benchmark for Decision-Critical Information State Tracking (DCIST), which makes the complete decision-critical event chain the prediction target (Figure~\ref{fig:overview}). Completeness is defined by the minimum events and relations required by the annotation protocol, rather than by every possible segmentation of speech. Evidence-supported refinements are allowed, but cannot compensate for missing required events. We instantiate the framework in financial dialogues using controlled synthetic speech, human-recorded scripted conversations, and paired variants that change speaker assignments or confirmation relations.

Our experiments address three questions: Can current systems recover full trajectories? Does this remain difficult when reference text and speaker labels are available? Do aggregate transcription errors reflect changes in trajectory quality? Our contributions are:
\begin{itemize}
\item A full-trajectory speech state tracking framework with a structured representation, minimum required chains, and a joint trajectory-level scoring protocol.
\item A Chinese financial dialogue dataset spanning two to four speakers, with synthetic and human-recorded speech, event provenance and temporal evidence, and paired variants.
\item A systematic evaluation of cascaded and direct-audio systems, reference-script controls, and targeted acoustic perturbations, exposing process-recovery failures not captured by transcription quality alone.
\end{itemize}

\section{Related Work}
\paragraph{Audio understanding and financial speech.}
AudioBench \citep{r1} provides multi-task audio evaluation; MMAU \citep{r2} covers auditory perception and reasoning; SpeakerSleuth \citep{r3} tests speaker consistency across dialogue turns. MultiFinBen \citep{r4} spans multilingual, multimodal financial applications, while FinAudio \citep{r5} targets financial transcription and summarization. These resources support content understanding. \bench{} further evaluates when information changes effectively, who causes the change, and within which scope it holds.

\paragraph{Attribution, events, and dialogue states.}
M3-SLU \citep{r7} and HEAR \citep{r8} evaluate speaker-attributed reasoning. SpeechEE \citep{r9} extracts spoken events and arguments. SpokenWOZ \citep{r6} evaluates turn-level slot-state maintenance, with related work studying synthetic speech and audio-based verification \citep{r12,r13}. \bench{} connects event actions and relations to speaker identity, scope, and time, evaluating their joint correctness over an entire required chain.

\paragraph{State evolution and memory.}
MultiWOZ \citep{multiwoz} annotates dialogue belief states; ProPara \citep{propara} tracks entity states in procedural text; OpenPI \citep{openpi} represents stepwise entity--attribute changes. The MEET consensus framework \citep{r10} preserves incremental propositions and decision histories with source and temporal metadata. We operationalize full decision-record recovery as a raw-audio prediction benchmark with structured gold trajectories and a joint metric. VoxMem \citep{r15} evaluates cross-session audio memory through information retrieval and temporal reasoning questions. Its retrieval target is question-conditioned; our target is the complete event chain itself. Appendix~\ref{app:comparison} contrasts these output requirements.

\section{Full-Trajectory State Tracking}
\subsection{Representation and Task}
Given an unsegmented multi-speaker recording and a task-wide public schema, a system predicts a set of trajectories. A trajectory is uniquely identified by an entity--field pair and contains temporally ordered decision cycles. Each cycle contains effective events recording an action, value, speaker, scope, predecessor, and evidence interval. Scope specifies the statement's authority and business effect, distinguishing, for example, application presentation, material verification, technical review, and formal approval. The schema provides 193 entity--field combinations, 64 scope labels, and permitted value structures. It does not reveal the fields, event counts, roles, or answers present in an individual sample.

An event requires an observable substantive change. Ordinary questions and unchanged repetitions do not automatically create nodes. Turns and events are not one-to-one: a turn can affect multiple fields, while the interpretation of one event can depend on preceding turns. Values record the information available at that point; a numeric candidate is not necessarily an approved result.

\Needspace{18\baselineskip}
\subsection{Events, Cycles, and Evidence}
\begin{table}[H]
\centering\small
\begin{tabularx}{\columnwidth}{@{}lXl@{}}
\toprule
Action & Trigger & State after \\
\midrule
\texttt{start} & First valid candidate or explicit initiation & active \\
\texttt{advance} & Substantive progress or resumption & active \\
\texttt{hold} & Actual suspension due to an explicit obstacle & held \\
\texttt{settle} & Explicit, source-supported scoped conclusion & settled \\
\bottomrule
\end{tabularx}
\caption{Event actions and resulting states.}\label{tab:actions}
\end{table}
Table~\label{call:tab:actions}\ref{tab:actions} defines four actions. The first observed event may be \texttt{hold} or \texttt{settle}; no unobserved \texttt{start} is invented. A successive \texttt{hold} is added only when the target value or a tracked critical condition materially changes. A revision that leaves processing suspended is \texttt{hold}, not \texttt{advance}. Settlement closes a cycle. An explicit reopening, revocation, or replacement decision starts another, which may consist of a single supported \texttt{settle} event.

Predecessor edges connect adjacent events within a cycle. Event and cycle identifiers may be renamed consistently. Speakers are numbered by first occurrence and evaluated without an optimal speaker permutation. Evidence intervals locate the principal triggering utterance; preceding context may justify its interpretation, but later conclusions cannot retroactively establish an earlier state.

\begin{figure*}[t]
\centering
\includegraphics[width=\textwidth]{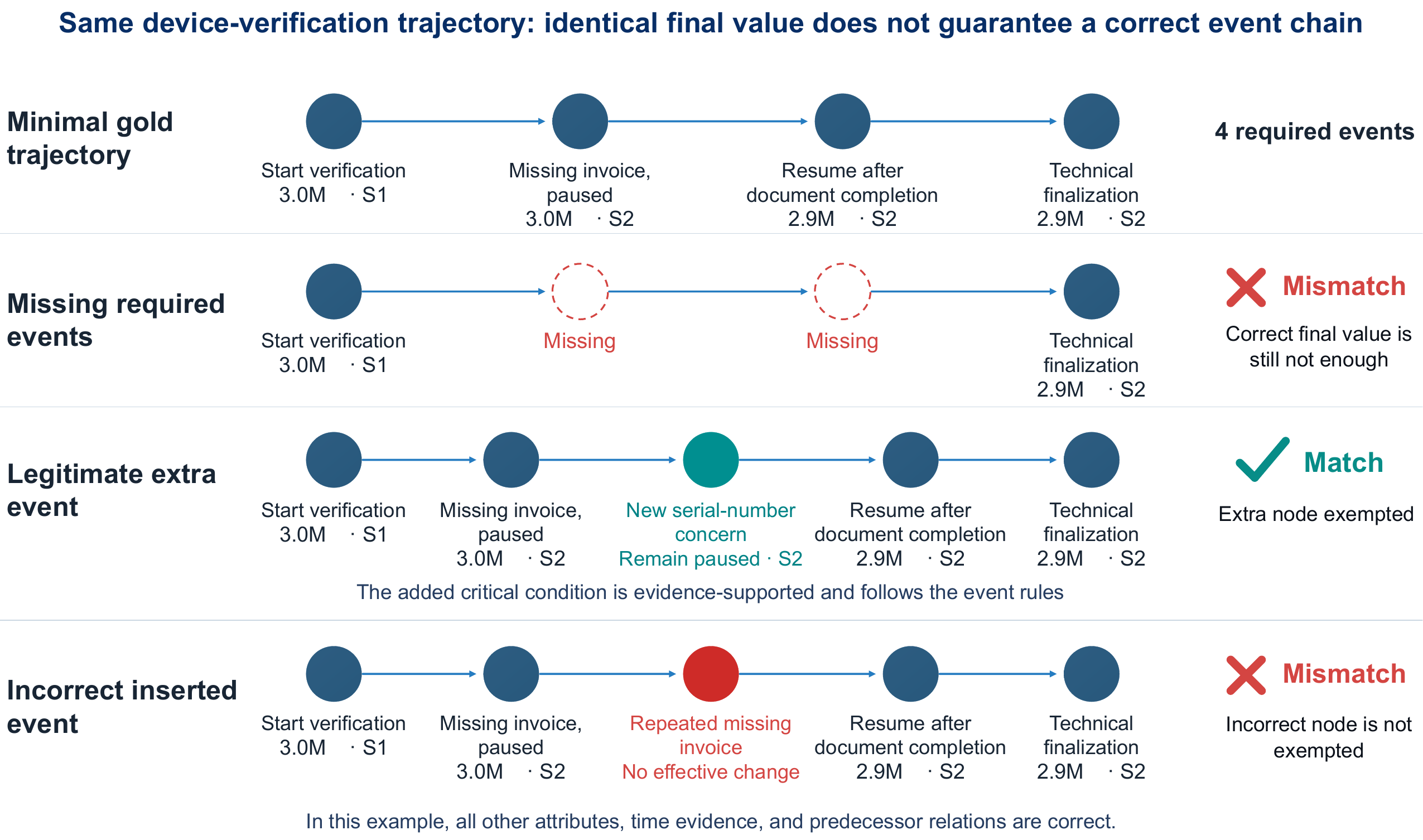}
\caption{Minimum required chains and additional nodes. S1 handles documentation and S2 performs technical verification. Omitting a required suspension or resumption fails the trajectory despite the correct terminal value. Supported refinements are exempt; an invalid insertion fails the trajectory. All other attributes, evidence intervals, and predecessor relations are assumed correct.}\label{fig:trajectory_matching}
\end{figure*}

\subsection{Matching and DCIST}
\paragraph{Required-chain matching.}
A prediction must cover every required event through an order-preserving, one-to-one alignment on the same tracked field. Matching checks cycle boundaries and order, action, value, speaker, scope, settlement result, predecessor relations, and time. One predicted node cannot satisfy two required nodes. The terminal state is derived from the last event rather than independently scored. Top-level explanatory text is not scored.

\paragraph{Expression equivalence.}
Free-text values admit many semantically equivalent expressions that cannot be exhaustively enumerated in the gold annotations. Strict surface matching may therefore reject a correct value because of a minor wording difference; under full-trajectory evaluation, this single mismatch can invalidate an otherwise correct trajectory. This motivates evaluating answer meaning rather than lexical overlap alone \citep{bulian-etal-2022-tomayto}. We therefore allow context-dependent semantic correspondences while preserving the model's freedom to express values. Astra first assesses whether field referents, scope meanings, and non-numeric value expressions convey the same information in the prediction, gold record, and relevant context. To reduce reliance on a single model's judgment, we introduce cross-model review \citep{verga2024juries}: Kimi-K3 examines the initial decision and rationale and challenges specific contradictions with the evidence or task rules. Candidates on which Astra and Kimi-K3 disagree are reviewed by a human, who makes the final equivalent or non-equivalent judgment. Approved mappings are sample-local and cannot supply missing facts, merge distinct propositions, or relax quantitative, action, speaker, authority, temporal, or structural constraints. Deterministic code performs trajectory matching and score aggregation using the frozen judgments.

\paragraph{Time and additional content.}
A predicted interval $P$ must have positive duration and satisfy $|P\cap G|/|P|\geq0.80$, where $G$ is the coarse gold interval. Independently reviewed, supported extra nodes are exempt from penalties. Reviewed extras whose evidence or rule boundary remains unresolved are also exempt; unreviewed candidates are not. Unsupported insertions, mislabeled required events, structural violations, and omissions remain errors. Exempt nodes are removed only in the scoring view, with predecessor edges contracted; this cannot repair an invalid edge or remove an extra cycle. Figure~\ref{fig:trajectory_matching} illustrates the distinction. Appendix~\ref{app:score} gives the full adjudication procedure and examples.

For recording $i$, let $G_i$ and $P_i$ be the gold and predicted trajectory counts, $L_i$ the exempt wholly new trajectories, and $TP_i$ the maximum one-to-one trajectory matching count:
\begin{align}
F_{1,i}&=\frac{2TP_i}{G_i+P_i-L_i},\\
\mathrm{DCIST}&=\frac{1}{N}\sum_{i=1}^{N}F_{1,i}.
\end{align}
Legitimate new trajectories are excluded from the predicted denominator but never add true positives. Exempt content cannot offset a required-event error. We separately report the proportion of unique exempt nodes among all parseable predicted nodes. Complete but malformed answers count as output errors; cohort and completion accounting are specified in Appendix~\ref{app:gemini}. Reference-script experiments disable time checks but retain other structural requirements.

\Needspace{18\baselineskip}
\section{Dataset}
We first design gold decision trajectories, then generate colloquial dialogue scripts conditioned on these trajectories, verify script--trajectory consistency, and finally produce audio with aligned event evidence. Synthetic speech enables controlled variation; human-recorded scripted conversations test the same task with natural human voices. Table~\label{call:tab:dataset}\ref{tab:dataset} summarizes the corpus. Synthetic dialogues cover 14 financial scenarios and human recordings cover 10, including credit due diligence, supply-chain finance, drawdown checks, audit adjustments, and insurance claims.

\begin{table}[H]
\centering\small
\begin{tabular}{@{}lrrr@{}}
\toprule
Subset & Audio & Hours & Tracks / events \\
\midrule
Synthetic & 2,246 & 100.39 & 5,319 / 18,227 \\
Human & 26 & 8.61 & 308 / 562 \\
Total & 2,272 & 109.01 & 5,627 / 18,789 \\
\bottomrule
\end{tabular}
\caption{Dataset size. Hours are playback duration; totals use unrounded values. A/B variants count as separate recordings.}\label{tab:dataset}
\end{table}

Synthetic recordings last 95.47--272.39 seconds; human recordings last 8.26--54.40 minutes. Human collection involved 22 participants. The estimated transcript-character rate, including pauses in the duration denominator, is 229.53 characters/minute (recording median: 230.77).

The synthetic subset contains 790 A/B pairs and 666 standalone scripts; the human subset contains 10 pairs and six standalone scripts. Paired variants change speakers, role assignments, or confirmation relations, with coherence checks on references and subsequent responses. They form 1,456 and 16 dialogue families, respectively; shared scripts and A/B variants remain in the same family. All recordings form fixed research evaluation cohorts, without a training/development partition. Across 58,396 synthetic--human script pairs, no normalized full-text duplicate was found; maximum character-five-gram Jaccard overlap is 0.50\%. Appendix~\ref{app:data} documents cohort membership, lexical audits, generation, consent, and alignment. Both subsets are scored with monetary equivalence, expression mappings, and minimum-chain matching.

\section{Experiments}
\subsection{Setup}
\paragraph{Systems.}
For human recordings, we evaluate FunASR \citep{funasr} with Astra, Kimi-K3, or Qwen3.8-Max, and VibeVoice-ASR \citep{vibeasr} with Kimi-K3 or Qwen3.8-Max. On synthetic speech, each frontend is paired with DeepSeek V4 Flash or Qwen3.8-Max. VibeVoice produces text, timestamps, and speaker labels. Neither receives manual transcripts, role answers, or sample-specific gold records. API text models use temperature 0, a 32,768-token output budget, and requests to disable thinking. Astra uses independent per-sample sessions with no prior sample or research-discussion history.

We additionally evaluate Gemini-4-Argon \citep{gemini4argon2026} and Gemini-3.8-Flash \citep{gemini38flash2026} directly on full-length human audio, without supplying a transcript, speaker count, role labels, or gold annotations. Both use the same audio-task query and time-grounded scoring protocol, with independent per-recording requests. Appendix~\ref{app:gemini} documents input encoding, request settings, and completion accounting.

Reference-script experiments provide all 26 scripts with or without speaker labels to Astra, Kimi-K3, DeepSeek V4 Flash, and Qwen3.8-Max, yielding 208 predictions. Turn order and text are unchanged; no additional cast list or speaker count is provided. Both conditions omit temporal verification.

\paragraph{Coverage and statistics.}
The original human-data experiments comprise 939 predictions across 41 model/input conditions. The main audio comparison extends the original five-pipeline cohort with two direct-audio systems, fixing the same 25 recordings, 293 gold trajectories, and 16 dialogue families for all seven systems. Each FunASR pipeline completes 26 recordings; each VibeVoice pipeline completes 25. Gemini retains the fixed denominator of 25. Other diagnostics retain their original systems and paired cohorts. All reported comparisons use frozen predictions and the same revised gold set.

DCIST is the sole primary metric. Terminal-value recall, A/B joint success, and failure categories are diagnostic. Perturbation experiments additionally report CER, fixed-tokenization WER, and concatenated minimum-permutation WER (cpWER), all relative to reading scripts. These may include performance deviations from the script and are not error rates against independently verified verbatim speech. Speaker identity remains strict in DCIST.

Reported human-data confidence intervals use 20,000 dialogue-family bootstrap samples, resampling A/B variants together while retaining recording-macro averaging. Two-sided paired sign-permutation tests flip differences at the family level, enumerating up to $2^{16}$ configurations. Holm correction jointly covers all 21 pairwise main-system comparisons; label ablations and B2 comparisons remain separate families of four and two tests. The threshold is 0.05. Appendix~\ref{app:statistics} reports the full methods and results.

\Needspace{22\baselineskip}
\subsection{Full-Trajectory Recovery}
\input{results_main.tex}
The seven systems score 5.73--21.35\% DCIST (Table~\label{call:tab:results}\ref{tab:results}). FunASR+Astra has the highest numerical score, recovering 82 of 293 complete trajectories. Direct-audio Gemini-4-Argon reaches 18.82\% with 72 matches, while Gemini-3.8-Flash reaches 6.59\% with 19. Kimi-K3 recovers 39 with VibeVoice and 38 with FunASR; Qwen recovers 18 with FunASR and 15 with VibeVoice. Since DCIST macro-averages recording-level F1, total matches alone do not determine ranking. Direct audio access therefore does not by itself ensure full-trajectory recovery.

The 95\% family-clustered interval for FunASR+Astra is [9.70, 34.62]. Across all 21 Holm-corrected comparisons, it exceeds same-frontend Kimi-K3 and Qwen by 11.55 and 15.59 percentage points (pp), respectively ($p=0.0498$ and $0.0371$), and Gemini-3.8-Flash by 14.75 pp ($p=0.0205$). Gemini-4-Argon exceeds FunASR+Qwen and Gemini-3.8-Flash ($p=0.0205$ and $0.0371$). The Astra--Argon difference of 2.53 pp is not significant ($p=0.7031$); no other pair passes the corrected 0.05 threshold. This does not establish equivalence between systems.

Multi-event recovery is especially difficult. The cohort contains 131 one-node, 103 two-node, and 59 three-or-more-node gold trajectories. FunASR+Astra recalls 42.75\%, 17.48\%, and 13.56\%, respectively; each Kimi pipeline matches only 1/59 in the last group. Speaker-count and duration strata are not monotonic, with scenario and chain complexity varying jointly (Appendix~\ref{app:additional}).

\Needspace{16\baselineskip}
\subsection{Synthetic Speech}
\input{results_synthetic.tex}
Four systems complete 8,957 outputs and share 2,224 recordings with 5,269 gold trajectories. Table~\label{call:tab:synthetic}\ref{tab:synthetic} applies the same monetary-equivalence, semantic-mapping, and minimum-chain rules. With time checks, FunASR+DeepSeek scores 32.71\%, followed by FunASR+Qwen (28.33\%), VibeVoice+DeepSeek (22.61\%), and VibeVoice+Qwen (21.64\%). Without time, VibeVoice+DeepSeek reaches 74.59\% and FunASR+DeepSeek scores 51.98\%. These results provide empirical evidence of task solvability: on short synthetic dialogues, systems can recover correct complete event chains from automatic transcripts without reference scripts. In particular, the higher untimed DCIST demonstrates successful joint recovery of content, attribution, and process structure, while timed evaluation additionally requires grounding those events in audio evidence. Appendix~\ref{app:synthetic_results} gives full coverage and scoring decomposition.

\Needspace{15\baselineskip}
\subsection{Reference Text and Speaker Labels}
\input{results_oracle.tex}
With labeled scripts and no time requirement, the highest DCIST remains 38.72\% (Astra, 133/308 complete matches; Table~\label{call:tab:oracle}\ref{tab:oracle}). Thus event selection, state interpretation, and relation recovery remain difficult even when designed wording and attribution are supplied. Removing labels lowers Astra, DeepSeek, and Qwen by 9.73, 6.73, and 5.10 pp, respectively; Kimi increases by 0.86 pp. Label benefits are significant for Astra and DeepSeek (both corrected $p=0.019$), but not Qwen or Kimi ($0.285$ and $0.766$).

\subsection{Terminal Values and Paired Dialogues}
We project each frozen prediction onto its terminal value after removing exempt refinements in the scoring view. Gold fields with null terminal values are excluded. This diagnostic retains field/value matching but not intermediate events, speaker, scope, or time; a terminal value can remain tentative or held rather than settled.

Across 266 eligible fields in the common audio cohort, terminal-value recall is 32.71--62.41\%, versus 5.26--30.45\% full-chain recall. FunASR+Astra recovers 166 terminal values, but 85 of those trajectories are incomplete or incorrect (51.20\%). The corresponding proportion for other pipelines is 71.53--85.26\%. With labeled scripts, Astra recovers 193 of 280 terminal values, of which 63 lack a correct full chain (32.64\%). Correct terminal content therefore does not establish process recovery. These are recall diagnostics, not DCIST F1; all denominators and intervals appear in Appendix~\ref{app:statistics}.

A/B joint success requires both variants of a shared entity--field pair to match their respective full gold trajectories. Audio systems share nine complete families and 87 field pairs; scripts contain ten families and 102 pairs. FunASR+Astra jointly recovers 21/87 (24.14\%), versus 41/102 (40.20\%) with labeled scripts. On the subset whose gold structure or attributes change after ignoring identifiers and evidence times, these counts are only 3/35 and 14/40. Other audio pipelines jointly recover 1.15--14.94\% of all paired fields. Role assignments and context can change together, so this is paired robustness, not a voice-only causal intervention.

\subsection{Scoring Sensitivity}
Kimi-K3 reviewed 2,366 deduplicated semantic candidates from the 939 human-data predictions. Keeping outputs, gold records, and other rules fixed, adopting all second-round objections and decisions on formerly unresolved candidates changes DCIST by at most 1.81 pp across all 41 conditions. The maximum occurs for Astra with unlabeled scripts: 28.99\% to 30.80\%, with complete matches increasing from 100 to 107. With labeled scripts, Astra increases from 38.72\% to 40.50\% (+1.78 pp): ten trajectories become matches and two cease to match, affecting 12/308 outcomes (3.90\%) with a net gain of eight. This indicates that the primary score is relatively stable under the judgment differences observed in sequential review. A single human reviewer subsequently conducted an independent, trajectory-by-trajectory review of all changed match outcomes. Per-trajectory score changes and corresponding adjudication records are retained for inspection. Temporal-threshold and speaker-ID sensitivity analyses appear in Appendix~\ref{app:additional}.

\subsection{Targeted Acoustic Perturbations}
We fix the trajectory model to Kimi-K3 and compare B2 keyword perturbation against unperturbed recordings using the same API route, prompt, and output budget. B2 selects 3,153 anchors for amounts, units, negation, confirmation, state expressions, and financial details, covering all 562 gold events. Selection uses original automatic word positions and coarse gold windows. Masks are frozen before downstream evaluation; approximately 4 dB noise changes 8.20\% of full-recording samples, preserving duration and all samples outside the masks.

\Needspace{21\baselineskip}
\input{results_noise_b2.tex}
In Table~\label{call:tab:noiseb2}\ref{tab:noiseb2}, FunASR's script-referenced WER rises by 1.99 pp and DCIST falls by 2.04 pp, from 9.68\% to 7.64\%. VibeVoice's WER rises by 1.47 pp, while DCIST falls by 3.32 pp, from 9.66\% to 6.34\%. Its smaller aggregate word-error increase accompanies a larger full-trajectory loss. The VibeVoice DCIST difference has a 95\% interval of [$-5.51,-0.96$] and corrected $p=0.039$; FunASR's interval is [$-8.15,2.40$], $p=0.625$. Only the former provides statistically detectable degradation under this test.

Historical A1 and A2 perturbations target speech outside gold evidence while protecting identity, authority, and critical premises, changing 7.32\% and 21.62\% of samples. Their non-monotonic results and automatic-speaker controls are reported separately in Appendix~\ref{app:noise}. Evidence-external speech is not necessarily semantically irrelevant. Differences in routes, transcription budgets, and noise duration preclude a direct causal comparison of A and B2 locations. Automatic word boundaries and retained redundant evidence also limit attribution of the B2 effect to individual words.

\subsection{Failure Analysis}
The five cascaded systems produce 1,465 gold-trajectory evaluation observations: 192 matches and 1,273 failures. Required-chain mismatch accounts for 807 failures (63.39\%), missing field candidates for 340 (26.71\%), and whole-output format errors for 112 (8.80\%). Thirteen predictions cover the required chain but contain invalid insertions; one violates trajectory structure. These are mutually exclusive matching barriers, not independent judgments of the model's underlying error. A missing candidate can also reflect field granularity, rather than absence of all corresponding information. This error census concerns the five cascades, not the direct-audio extension. Appendix~\ref{app:astra_script_errors} provides a detailed discussion of Astra's error distributions under both reference-script conditions, including category definitions and counts over all unmatched trajectories.

Exempt extras constitute 10.50--26.37\% of parseable nodes for the audio pipelines and 17.58--27.28\% for labeled scripts. This measures output granularity, not hallucination rate, and does not increase true positives. Taken together, terminal-value projections, paired variants, reference-text controls, and acoustic perturbations show why full-record correctness must be evaluated directly rather than inferred from a local extraction or transcription score.

\section{Conclusion}
We introduce \bench{}, a speech-based benchmark for event-chain tracking that jointly evaluates decision evolution across speakers, information attribution, scope, and temporal evidence. It provides the first quantitative assessment in speech understanding of how correct terminal values can mask errors in reconstructing the underlying event history. The benchmark offers a process-oriented evaluation of speech understanding for settings requiring accurate event reconstruction from audio, including financial operations, accident investigations, and critical meetings.

Our experiments show that accurately recovering complete event histories remains challenging for current models. The highest DCIST score on human recordings is 21.35\%. For the best-performing system, FunASR+Astra, 51.20\% of the correctly recovered terminal values still correspond to incomplete or incorrect trajectories. Acoustic perturbation experiments further show that overall transcription error rates do not adequately reflect the recovery of decision-critical information: a higher overall transcription error rate does not necessarily imply a lower DCIST score. Complete decision-process recovery must therefore be evaluated directly, rather than inferred from transcription accuracy or terminal-value correctness.

We hope that DCIST will support the development of next-generation speech agents.

\FloatBarrier
\label{main:end}
\input{limitations_ethics.tex}
\par
\Needspace{12\baselineskip}
\begin{samepage}
\section*{Acknowledgments}
We thank Xudong Cai, Han Cui, Huifang Du, \mbox{Jiajing Gong}, Hongjin Hu, Tong Hu, Jia Kang, Yifei Sun, Xiaoyu Yan, Kai Yang, and Yuting Zhang for their contributions to the human-performed recordings in DecisionTraceBench. We also thank the participants who requested anonymous acknowledgment for their time and contributions to the recording sessions.
\par
\end{samepage}

\bibliography{references}
\clearpage
\appendix
\input{appendices.tex}

\end{document}

%% file: results_main.tex
\begin{table}[H]
\centering
\fontsize{9.5}{11.5}\selectfont
\setlength{\tabcolsep}{3pt}
\begin{tabular}{@{}>{\raggedright\arraybackslash}p{0.22\columnwidth}>{\raggedright\arraybackslash}p{0.34\columnwidth}rrr@{}}
\toprule
\shortstack[l]{Input\\pipeline} & \shortstack[l]{Trajectory\\model} & N & DCIST & TP \\
\midrule
FunASR & Astra & 25 & 21.35 & 82 \\
FunASR & Kimi-K3 & 25 & 9.80 & 38 \\
FunASR & Qwen3.8-Max & 25 & 5.75 & 18 \\
VibeVoice & Kimi-K3 & 25 & 11.09 & 39 \\
VibeVoice & Qwen3.8-Max & 25 & 5.73 & 15 \\
\midrule
Direct audio & Gemini-4-Argon & 25 & 18.82 & 72 \\
Direct audio & Gemini-3.8-Flash & 25 & 6.59 & 19 \\
\bottomrule
\end{tabular}
\caption{Full-trajectory recovery (DCIST, \%) on a fixed cohort of 25 unperturbed human recordings, 16 dialogue families, and 293 gold trajectories. TP counts complete matches. VibeVoice uses native full-length transcription; Gemini receives full-length audio directly. All systems use the same minimum-chain protocol and time threshold. Appendix~\ref{app:gemini} specifies the direct-audio configuration; Appendix~\ref{app:statistics} reports all 21 Holm-corrected comparisons and the retained cascade intervals.}\label{tab:results}
\end{table}

%% file: results_synthetic.tex
\begin{table}[H]
\centering\small
\begin{tabular}{@{}llrr@{}}
\toprule
ASR & Reasoner & DCIST & No time \\
\midrule
FunASR & DeepSeek & 32.71 & 51.98 \\
FunASR & Qwen & 28.33 & 46.47 \\
VibeVoice & DeepSeek & 22.61 & 74.59 \\
VibeVoice & Qwen & 21.64 & 67.70 \\
\bottomrule
\end{tabular}
\caption{Synthetic results on 2,224 shared recordings (5,269 gold trajectories). Scores are percentages. The no-time column disables temporal checks only.}\label{tab:synthetic}
\end{table}

%% file: results_oracle.tex
\begin{table}[H]
\centering
\fontsize{9.5}{11.5}\selectfont
\setlength{\tabcolsep}{2pt}
\begin{tabular}{@{}lrrrrr@{}}
\toprule
Model & Labeled & TP & Unlabeled & TP & $\Delta$ (pp) \\
\midrule
Astra & 38.72 & 133 & 28.99 & 100 & +9.73 \\
Kimi-K3 & 22.72 & 78 & 23.58 & 77 & -0.86 \\
DeepSeek V4 Flash & 14.04 & 41 & 7.31 & 18 & +6.73 \\
Qwen3.8-Max & 18.89 & 55 & 13.79 & 40 & +5.10 \\
\bottomrule
\end{tabular}
\caption{Reference-script results: 26 scripts, two input conditions, four models, and 208 predictions. Scores are percentages; $\Delta$ is labeled minus unlabeled in percentage points. Both conditions disable time checks but retain strict speaker matching. TP counts complete trajectories.}\label{tab:oracle}
\end{table}

%% file: results_noise_b2.tex
\begin{table}[H]
\centering
\fontsize{10}{12}\selectfont
\setlength{\tabcolsep}{4pt}
\begin{tabular}{@{}lrrrr@{}}
\toprule
Condition & CER & WER & cpWER & DCIST \\
\midrule
\multicolumn{5}{@{}l}{\textit{FunASR ($N=26$)}} \\
Original & 17.12 & 21.69 & 31.18 & 9.68 \\
B2 & 18.77 & 23.68 & 33.15 & 7.64 \\
\midrule
\multicolumn{5}{@{}l}{\textit{VibeVoice ($N=22$)}} \\
Original & 9.01 & 11.10 & 13.29 & 9.66 \\
B2 & 10.41 & 12.57 & 14.21 & 6.34 \\
\bottomrule
\end{tabular}
\caption{Same-route Kimi-K3 comparison with B2 keyword noise. All metrics are percentages. FunASR and VibeVoice share 26 and 22 completed recordings across conditions, respectively. Within each frontend, the API route, prompt, and output budget are held constant. Transcription metrics use the reading scripts as references.}\label{tab:noiseb2}
\end{table}

%% file: limitations_ethics.tex
\section*{Limitations}
\paragraph{Representativeness.}
The human subset consists of 26 scripted Chinese recordings, with limited coverage of spontaneous meetings or multilingual practice. A/B variants share content, so the number of independent examples is smaller than the recording count.

\paragraph{Representation and scoring.}
Minimum required chains do not enumerate every valid event. Once the required chain is correctly recovered, additional accurate, fine-grained events do not increase the primary score, limiting the metric's ability to distinguish such incremental capabilities.

\paragraph{Temporal localization.}
The coarse-window overlap rule tests evidence placement more directly than word-boundary precision. The small score changes under 50--90\% predicted-window overlap thresholds do not imply accurate boundaries. Requiring coverage of half the gold window sharply reduces scores, showing that placement and boundary coverage are different targets. Fine-grained temporal evaluation requires finer annotation.

\paragraph{Comparability and generalization.}
Native full-length processing, chunked inference, and auxiliary speaker alignment involve different computational and information conditions. Incomplete coverage can also affect comparisons. Family-clustered intervals quantify variation within the collected dialogue set, not repeated-generation variance or participant-independent population variation. Both subsets are fixed research evaluation cohorts. Low cross-source lexical overlap does not establish semantic-template independence. Validation in other record-sensitive settings requires independent data and experiments.

\section*{Ethical Considerations}
All human-recording participants provided informed consent, documented through signed forms and supplementary voice records, covering recording, foundational research, model evaluation, and public release. Attribution preferences are recorded separately. Dialogues are fictional and do not request real client or personal financial information. Participants were informed about transcription, annotation, segmentation, mixing, and noise augmentation, and that their voices may remain identifiable.

Public release follows each participant's permissions and uses anonymous speaker identifiers. Human recordings authorized for public release are distributed under the Creative Commons Attribution-NonCommercial 4.0 International License (CC BY-NC 4.0); commercial use requires separate permission. Identity-to-recording mappings are not public, and acknowledgments follow individual preferences. Participants may withdraw permission for future use; affected recordings will be excluded from subsequent official versions, although previously distributed third-party copies cannot be guaranteed to be recalled.

Speech synthesis uses Alibaba's open-source CosyVoice 2 model \citep{cosyvoice2}. Reference-voice resources include public AISHELL corpora \citep{aishell1,aishell3} and author-provided recordings authorized for research synthesis; the 100.39-hour batch reported here is linked to AISHELL-1/3 entries. Third-party audio and derivatives are managed under their respective source licenses, separately from the human-recording release. Volunteer recording consent is not treated as voice-cloning authorization. The benchmark evaluates research systems rather than establishing their suitability for operational lending or other high-stakes decisions.

\paragraph{AI assistance.}
Generative models assisted script construction, semantic adjudication, code preparation, and manuscript drafting, translation, and language editing. The model roles relevant to the experiments are described in the methods and appendices. Responsibility for the methods, results, citations, code, and claims remains with the authors.

%% file: appendices.tex
\section{Worked Scoring Examples}\label{app:examples}
\subsection{A Four-Event Required Chain}
Table~\ref{tab:example} gives a constructed example, not an observed dataset result. S1 handles materials and S2 performs technical verification. S2's technical conclusion does not approve a loan. Speakers are numbered in first-occurrence order. The value at an event is the current candidate or recorded result, not necessarily a settled value.

\begin{table}[t]
\centering\small
\begin{tabularx}{\columnwidth}{@{}Xl@{}}
\toprule
Utterance (English translation) & Event \\
\midrule
S1: Start checking this equipment at CNY 3 million. & start: 3.0M \\
S2: An invoice is missing. Pause this; we cannot finalize it yet. & hold: 3.0M \\
S2: The invoice and serial numbers now check out. Depreciation gives 2.9 million. Resume verification. & advance: 2.9M \\
S2: Yes, record the equipment at 2.9 million. & settle: 2.9M \\
\bottomrule
\end{tabularx}
\caption{A required chain in an established technical-verification context. M denotes million CNY.}\label{tab:example}
\end{table}

The following is one trajectory with one cycle and four events. Intervals illustrate millisecond evidence fields. JSON Unicode escapes preserve the original Chinese monetary units: \texttt{\textbackslash u4e07} denotes ten thousand; the additional \texttt{\textbackslash u5143} denotes yuan. Thus number 300 with the first unit is CNY 3 million. No gold or prediction value is altered by the display encoding.

\begin{Verbatim}
{"tracks": [{
  "entity_id": "collateral_equipment",
  "field_id": "technical_recognized_value",
  "derived_final_state": "settled",
  "cycles": [{"cycle_id": "C1",
    "events": [
    {"event_id":"E1", "action":"start",
     "speaker_id":"S1",
     "scope_id":"material_verification",
     "value":{"number":"300","unit":"\u4e07"},
     "source_event_id":null,
     "start_ms":10000, "end_ms":15000},
    {"event_id":"E2", "action":"hold",
     "speaker_id":"S2",
     "scope_id":"technical_review",
     "value":{"number":"300","unit":"\u4e07"},
     "source_event_id":"E1",
     "start_ms":18000, "end_ms":24000},
    {"event_id":"E3", "action":"advance",
     "speaker_id":"S2",
     "scope_id":"technical_review",
     "value":{"number":"290","unit":"\u4e07"},
     "source_event_id":"E2",
     "start_ms":55000, "end_ms":63000},
    {"event_id":"E4", "action":"settle",
     "speaker_id":"S2",
     "scope_id":"technical_review",
     "value":{"number":"290","unit":"\u4e07"},
     "result":"accepted",
     "source_event_id":"E3",
     "start_ms":68000, "end_ms":72000}
    ]}]
}]}
\end{Verbatim}

E2 inherits the 3-million candidate; it is not an accepted technical value. Renaming E1--E4 to P1--P4 while consistently renaming references preserves the structure. Omitting an event, changing the speaker, or substituting a different scope does not.

\subsection{Correct Terminal Value, Incorrect Chain}
The following prediction contains only the first and final events. Its terminal amount is correct and its own predecessor edge is well formed, but it omits the required suspension and resumption.

\begin{Verbatim}
{"tracks": [{
  "entity_id": "collateral_equipment",
  "field_id": "technical_recognized_value",
  "derived_final_state": "settled",
  "cycles": [{"cycle_id": "C1",
    "events": [
    {"event_id":"P1", "action":"start",
     "speaker_id":"S1",
     "scope_id":"material_verification",
     "value":{"number":"300","unit":"\u4e07\u5143"},
     "source_event_id":null,
     "start_ms":10000, "end_ms":15000},
    {"event_id":"P2", "action":"settle",
     "speaker_id":"S2",
     "scope_id":"technical_review",
     "value":{"number":"290","unit":"\u4e07"},
     "result":"accepted",
     "source_event_id":"P1",
     "start_ms":68000, "end_ms":72000}
    ]}]
}]}
\end{Verbatim}

Here $G=P=1$, $L=0$, and $TP=0$, so F1 is zero. If a recording instead contains two gold trajectories and two predictions, with the other trajectory fully correct, then $TP=1$ and F1 is 0.50. Node count alone is not the criterion: a longer prediction must be checked for legitimate refinements, whereas a missing required node always prevents a match.

For the gold interval [68000,72000] ms, prediction [67500,72500] overlaps by 4000 ms, or 0.80 of predicted duration, and passes. Prediction [66000,74000] overlaps by 0.50 and fails. Passing the time rule does not substitute for correct event attributes or relations.

\subsection{Additional Nodes and Trajectories}
\paragraph{Legitimate refinement.}
Between E2 and E3, suppose S2 states at [30000,34000] ms: ``The serial number also disagrees with the list. That must be resolved too; 3 million is still the candidate, and verification remains paused.'' This new critical obstacle supports another \texttt{hold} node X. Its predecessor is E2, and E3 points to X. After adjudication, only the scoring view contracts E2--X--E3 into E2--E3. All required nodes are covered, F1 is 1, and one of five predicted nodes is exempt (20\%).

\paragraph{Invalid insertion.}
Insert the same \texttt{hold} node, but let its evidence merely repeat: ``The same invoice is still missing. Keep waiting.'' Neither the target value nor a critical condition changes. The insertion is invalid even though all four required nodes are present: $TP=0$ and F1 is zero. No extra node is exempt. Deleting this error to award full credit is not permitted.

\paragraph{A wholly new supported trajectory.}
Instead, retain the four-node chain and add S1's statement at [76000,80000] ms: ``We are also applying for a five-million loan; submit that amount for now.'' It supports a separate \texttt{loan\_request.application\_amount} trajectory with one \texttt{start}, scope \texttt{application\_presentation}, a five-million value, and a null predecessor. This field is absent from the example gold. If all output content is supported, then $G=1$, $P=2$, $L=1$, and $TP=1$: F1 remains 1, not greater. Its node contributes to the extra proportion (one of five). An additional unsupported approval would prevent whole-trajectory exemption, although the supported start could still count separately in the auxiliary node statistic. These examples construct predictions for illustration; scoring never edits model answers to create them.

\section{Relation to Existing Tasks}\label{app:comparison}
Full-trajectory recovery evaluates states together with the effective events that produce them. Financial fields instantiate the representation; the structural target is not intrinsically financial.

\begin{table*}[t]
\centering\small
\begin{tabularx}{\textwidth}{@{}p{.22\textwidth}XX@{}}
\toprule
Task family & Existing evaluation or representation & Full-trajectory extension \\
\midrule
MultiWOZ, ProPara, OpenPI \citep{multiwoz,propara,openpi} & Dialogue belief states or entity changes in procedural text & Joint recovery of spoken evolution, source identity, scope, and acoustic evidence \\
SpokenWOZ and spoken DST \citep{r6,r12,r13} & Turn-level slot updates, including synthetic-speech and audio-verification studies & Event actions, authority scope, and time evidence evaluated over the full chain \\
MEET consensus model \citep{r10} & Incremental decision history with provenance and time metadata & A unified raw-audio prediction benchmark with joint trajectory scoring \\
M3-SLU / HEAR \citep{r7,r8} & Speaker binding, provenance questions, and voice-swap controls & Attribution as part of effective decision events and their scoped consequences \\
SpeechEE \citep{r9} & Spoken event types, triggers, and arguments & The continuous suspension, resumption, and resolution of a proposition \\
Audio Arena \citep{r14} & Correction, memory, and state maintenance in interactive speech & Traceable record recovery from an existing multi-speaker conversation \\
VoxMem \citep{r15} & Question-conditioned cross-session retrieval and temporal reasoning & The complete event chain, rather than question-selected information, as the output \\
\bottomrule
\end{tabularx}
\caption{Structured comparison of prediction targets.}\label{tab:comparison}
\end{table*}

\section{Data and Implementation}\label{app:data}
\subsection{Distribution and Scenarios}
\begin{table*}[t]
\centering\small
\begin{tabular}{@{}lrr@{}}
\toprule
Statistic & Synthetic & Human-recorded \\
\midrule
Scenario categories & 14 & 10 \\
Two / three / four speakers & 448 / 1,357 / 441 & 5 / 14 / 7 \\
A/B pairs (recordings) & 790 (1,580) & 10 (20) \\
Standalone scripts & 666 & 6 \\
Dialogue families & 1,456 & 16 \\
Minimum / median / maximum duration (s) & 95.47 / 154.50 / 272.39 & 495.50 / 997.53 / 3,264.00 \\
Mean duration (s) & 160.92 & 1,192.62 \\
\bottomrule
\end{tabular}
\caption{Recording distribution. Speaker counts refer to script roles. Both A/B variants are counted as audio, but grouped into one family.}\label{tab:distribution}
\end{table*}

Synthetic scenarios comprise working-capital review (235 recordings), equipment valuation (209), guarantee review (208), use-of-funds investigation (200), drawdown-condition verification (194), repayment reconciliation (190), supply-chain financing (180), invoice financing (172), project drawdowns (170), certificate-of-deposit pledges (167), receivables assignment (91), syndicated participation intentions (91), mortgage negotiation (76), and export proceeds (63).

The human subset contains three recordings each for corporate credit due diligence, equipment-renewal loan due diligence, resale-home mortgages and transaction funding, post-approval drawdown conditions, working-capital credit committees, and property-insurance claims. Audit adjustments, supply-chain financing, financial leasing/sale-and-leaseback, and syndicated loan share coordination each contribute two.

\subsection{Voice Sources, Evidence, and Speaking Rate}
The 100.39-hour synthetic batch uses 193 source-speaker entries: 94 from AISHELL-1 \citep{aishell1} and 99 from AISHELL-3 \citep{aishell3}. Reference provenance includes the source corpus, speaker identifier, file location, and hash. These are source-corpus entries, not a cross-corpus deduplication of physical identities. Turn-level synthesis and concatenation provide exact assembly positions. All 3,538 human-script turns have start/end times; trajectory events reference corresponding coarse windows on the original recording timeline.

The authors attended and supervised the human recording sessions, checking decision-critical content and the corresponding speaker roles against the recording scripts. This supervision ensured that the recorded key information was consistent with the events underlying the trajectory gold. Aligned scripts provide temporal locations for event evidence on the original recording timeline; trajectory annotation relies on supervised collection and event-level evidence rather than exhaustive word-level annotation. Candidate adjudication preserves differences between actual model input and reference scripts. Collection dates or scripted settings cannot establish relative years in an utterance, and scripted values cannot replace values in the actual transcription.

Human collection involved 22 participants, distinct from the two-to-four role labels in each script. Unperturbed FunASR transcripts contain 118,620 Chinese characters, Latin letters, and digits over 31,008.216 seconds. After excluding punctuation and whitespace but retaining repetitions and fillers, the overall rate is 229.53 characters/minute. Per-recording mean, median, and range are 231.53, 230.77, and 198.46--275.48. Duration includes pauses; these are automatic transcript-character estimates rather than articulation rates.

\subsection{Generation and Inference}
\begin{table*}[t]
\centering\small
\begin{tabularx}{\textwidth}{@{}p{.20\textwidth}p{.26\textwidth}X@{}}
\toprule
Stage & Model / implementation & Configuration \\
\midrule
Script generation & DeepSeek V4 Flash & Temperature 0.45; generates colloquial scripts conditioned on predesigned gold trajectories \\
Cross-review & GLM5.3-Flash; Qwen3.8-Max & Temperature 0.05; respective output budgets 18,000 and 9,000 tokens \\
Speech synthesis & CosyVoice2-0.5B \citep{cosyvoice2} & 24 kHz, mono, 16-bit; speed 1.0; half precision; turn-level synthesis \\
Full-length transcription & FunASR & Punctuation-based segmentation; processing batch configured for 300 s; no supplied speaker count \\
Native transcription & VibeVoice-ASR & Non-sampling decoding, SDPA; 32,768-token budget for synthetic, original human, and B2; 65,536 for historical A1 \\
Chunked transcription & VibeVoice with FunASR speaker alignment & Continuous non-overlapping 300-s and 60-s variants; 16,384 tokens per chunk; uncertain identities retained \\
API trajectory recovery & Kimi-K3; Qwen3.8-Max; DeepSeek V4 Flash & Temperature 0; 32,768-token budget; requests disable thinking; actual responses and usage retained \\
Independent-session recovery & Astra (gpt-6-astra) & One sample per session with no other conversation history \\
\bottomrule
\end{tabularx}
\caption{Generation and inference settings. Different frontends, chunk lengths, and auxiliary inputs define distinct conditions.}\label{tab:config}
\end{table*}

The recorded script-experiment request identifiers are \texttt{kimi-k3}, \texttt{qwen3.8-max}, \texttt{deepseek-v4-flash}, and \texttt{gpt-6-astra}. The speech frontend methods are described by \citet{funasr,vibeasr}.

Construction proceeds through gold-trajectory design, gold-conditioned colloquial script generation, multi-model cross-review, rule checks, recording production, and time alignment. All 1,456 admitted synthetic script groups have both GLM and Qwen review records; 14 also have Kimi-K3 review, totaling 2,926 archived admission reviews. Hash-linked audio, text, gold records, and timing metadata support provenance. Human gold review covers all 308 trajectories and 3,538 scripted turns, checking textual support, substantive state changes, and settlement evidence.

\subsection{Direct-Audio Gemini Evaluation}\label{app:gemini}
The requested and returned service identifiers are \texttt{gemini-4-argon} and \texttt{gemini-3.8-flash}. Each independent request includes one complete MP3 recording and the frozen audio-task query, without a transcript, cast list, speaker count, gold labels, or previous conversation. Temperature is 0, streaming is disabled, and provider-default thinking is retained. The initial output allowance is 15,360 tokens. Flash receives one 32,768-token retry per originally truncated request for R0005, R0006, and R0011, using the same complete audio and query; these are fresh requests rather than concatenated answers. Existing completed answers are retained regardless of their correctness.

Eight recordings are encoded as full-length mono 48-kbps MP3 copies to meet request-body limits; both systems receive identical audio bytes for these recordings. No recording is cropped or accelerated. The remaining inputs retain their original MP3 bytes. Input hashes, conversion receipts, request settings, and unmodified responses are retained for reproduction. These encoding and budget differences are part of the direct-audio configuration, not matched-compute ablations against the cascades.

The main comparison fixes the existing 25-recording cohort and excludes R0024 for every system. Each direct-audio recording contributes one F1 value; an absent complete prediction contributes zero without changing $N=25$. Argon returns 25 complete answers; Flash returns 24, including one format-invalid answer, and has one absent answer. These completion counts are distinct from semantic error categories. Earlier cascade, synthetic, and perturbation analyses retain their reported completed-output cohorts.

The same monetary-equivalence, field/scope/value review, additional-node adjudication, speaker, structure, and 80\% temporal-overlap rules apply. Astra's candidate judgments are checked by Kimi-K3 in 48 independent per-output review batches covering 1,471 decisions: 451 expression candidates, 878 node candidates, and 142 whole-trajectory candidates. Eight objections are examined against the frozen rules; the finalized review leaves all trajectory-match outcomes and both main DCIST scores unchanged. This is sequential review with access to the first judgments, not blind inter-model agreement. The earlier 939-prediction sensitivity analysis remains a separate experiment.

\subsection{Pairs, Evaluation Cohorts, and Gold Revisions}
Counterfactual pairs vary speakers, responsibilities, or confirmation relations around the same critical information. Pair review checks references and later responses; residual contextual differences are retained rather than represented as a pure voice swap. Families group shared scripts and A/B variants. The cohort manifest assigns all 2,246 synthetic recordings and all 26 human recordings to fixed research evaluation cohorts, with no training or development partition. Our evaluation assesses existing systems rather than train-to-test generalization; family identifiers preserve paired dependence in analyses.

A lexical overlap audit concatenates spoken text in script order, removes punctuation, whitespace, and speaker labels, retains Chinese characters, ASCII letters and digits, and lowercases letters. Across 315 cross-family human-recording pairs, there are no normalized whole-script duplicates. Character-five-gram set Jaccard overlap has median 0.08\%, 90th percentile 0.41\%, and maximum 1.65\%. No identical complete turn of at least 12 characters is shared across human families. Within-family A/B overlap is intentional.

Across all 58,396 synthetic--human script pairs, normalized whole-script duplicates and identical complete turns of at least 12 characters are both absent. Character-five-gram Jaccard overlap has median 0.018\%, 90th percentile 0.077\%, and maximum 0.497\%. No normalized whole-script duplicate occurs between distinct synthetic families. These lexical checks do not establish semantic-template independence, speaker independence, or absence of pretraining contamination.

The project repository is \url{https://github.com/QFIN-tech/DecisionTraceBench}. Appendix~\ref{app:examples} provides constructed scoring examples. The complete audio dataset and gold annotations will be publicly released after acceptance.

The current human gold version corrects unsupported monetary values and a repetition by an unauthorized participant previously treated as suspension. Three files, four trajectories, and six events are affected; total trajectory and event counts are unchanged. Triggering turns, speakers, conclusions, and time windows are synchronized; other disputed boundaries remain unchanged. Replacing only these records leaves DCIST unchanged for all 939 predictions, because other required constraints still prevent full matches.

\subsection{Noise Verification}
A1, A2, and B2 each contain 26 full-length copies. Samplewise checks confirm unchanged samples outside masks and unchanged protected identity, authority, premise, and gold regions for A conditions. A2 changes 21.62\% overall; 25 recordings exceed 20\%, while one reaches 11.58\% because its safe evidence-external region is insufficient. B2 changes 40,659,118 samples (8.20\%), with all 26 passing input-correspondence checks. FunASR completes 26 B2 transcriptions; VibeVoice completes 22. The four missing VibeVoice samples are not measured zeros.

\subsection{Synthetic Evaluation}\label{app:synthetic_results}
\input{results_synthetic_all.tex}
The four systems complete 2,232, 2,245, 2,239, and 2,241 recordings (99.38\%, 99.96\%, 99.69\%, and 99.78\% of 2,246). The shared cohort contains 2,224 recordings and 5,269 gold trajectories; full-coverage scores appear in Table~\ref{tab:synthetic_all}. Existing predictions and gold remain frozen. Actions, speaker identity, scope, numerical content, cycles, and predecessor relations remain required. Primary semantic review comprises 18,518 candidate decisions. Review closure covers every candidate capable of changing DCIST; a candidate can be skipped only when code establishes score invariance, never by marking an unreviewed item equivalent or unresolved. This procedure determines the main score but does not classify every superfluous node, so we do not report a complete synthetic extra-node ratio.

Required events use frozen synthesis timestamps. Extra-event semantic and structural decisions are shared across the timed and untimed conditions; timed scoring additionally checks the principal triggering turn against synthesis metadata at the 80\% overlap threshold. Removing time removes that check only. The scoring implementation passes 24 regression tests and the final evidence/timing audits.

\input{results_synthetic_decomposition.tex}
Table~\ref{tab:synthetic_decomposition} separates successive scoring changes on fixed outputs. Under temporal constraints, monetary normalization adds 0.02--0.28 pp, expression mappings add 11.21--21.35 pp, and extra-node adjudication adds 0.05--0.81 pp. These increments depend on the order of operations, not on improved model predictions. The large expression component highlights the sensitivity of literal field/scope/value matching; it is distinct from the 1.81-pp second-model sensitivity bound measured on human experiments.

\section{Acoustic Perturbations}\label{app:noise}
\begin{table*}[t]
\centering\small
\begin{tabular}{@{}llll@{}}
\toprule
Condition & Region & Mask duration & Changed samples \\
\midrule
Original & No added noise & --- & Reference \\
A1 & Outside protected gold evidence & 2,270.24 s & 7.32\% overall \\
A2 & Expanded evidence-external speech & 6,705.04 s & 21.62\%; one 11.58\% exception \\
B2 & Critical short words & 2,541.69 s & 8.20\% overall \\
\bottomrule
\end{tabular}
\caption{Frozen perturbation conditions, each with 26 full-length recordings and approximately 4 dB noise. Changed-sample proportions are measured against original waveforms. The other 25 A2 recordings range from 21.01\% to 22.89\%.}\label{tab:noise}
\end{table*}

B2 uses 3,153 anchors covering 562 gold events: 1,146 numeric/unit/term/date anchors, 925 negation/state/confirmation anchors, and 1,082 financial-detail anchors. Each is expanded by 160 ms on both sides, with overlapping intervals merged. Boundaries are automatic, not manually verified words. Redundant evidence outside masks remains audible; no mask is selected or changed to obtain a desired downstream error.

\subsection{Historical Evidence-External Noise}
\input{results_noise_history.tex}
In Table~\ref{tab:noisehistory}, FunASR's script-referenced WER increases from 21.69\% to 24.17\% (A1) and 29.51\% (A2), while DCIST is 12.41\%, 7.91\%, and 10.19\%, respectively. Larger global deviation in A2 does not yield greater trajectory loss: the decline is 2.22 pp versus 4.50 pp for A1. On 23 shared VibeVoice recordings, WER increases from 11.13\% to 12.34\% but DCIST increases from 10.10\% to 11.22\%. These non-monotonic patterns motivate separating recognition changes from downstream variation through repeated inference and event-level analysis. They are not a controlled causal contrast with B2.

\subsection{Automatic Speaker Control}
\input{results_speaker_noise.tex}
Replacing noisy automatic attribution with the original recording's automatic segmentation increases Kimi's score by 1.82 pp for FunASR A1 and 1.24 pp for VibeVoice A1, but decreases FunASR A2 by 1.25 pp. These controls do not consistently eliminate trajectory error. They use automatic identities, not human gold speakers, and are not substitutes for diarization error evaluation. None of the three corrected tests is significant (Table~\ref{tab:noise_tests}).

\section{Scoring Protocol}\label{app:score}
\subsection{Required Attributes and Relations}
Matching jointly checks the tracked field, cycles, and event attributes. Cycle count, boundaries, and order must agree; an extra cycle cannot be deleted as an exempt node. Within each cycle, every required event must be covered one-to-one in order. Action, speaker, and settlement result must agree. Event and cycle identifiers can be consistently renamed, but speaker permutation and content editing cannot repair a prediction. The derived final state and top-level reason are not separately scored; content nested inside a value belongs to that value.

Numbers are output as Arabic-digit strings with spoken units. Monetary equivalence uses exact decimal conversion within an explicit currency and unit scale; it neither guesses missing magnitudes nor applies exchange rates or general duration conversions. Composite values must preserve all critical components, with unordered object keys. A node that expresses a required event incorrectly is a mislabeled substitution, not a legitimate extra; exemption cannot bypass required-event constraints.

\subsection{Sequential Semantic Review}\label{app:semantic_review}
\paragraph{First stage.}
Astra judges whether gold and predicted expressions are equivalent in context, saving evidence and rationale; insufficient evidence remains unresolved. Field mappings require the same subject, attribute, and business proposition. Scope mappings require the same authority conditions, confirmation level, and business effect. Non-numeric values allow synonyms, cross-language terms, abbreviations, or lossless structural representations. Equal numbers alone do not justify merging fields or scopes. Amounts, durations, negation, time ranges, and necessary conditions cannot be deleted or supplied.

\paragraph{Second stage.}
We simplify this stage to a directed check of explicit first-round contradictions. Primary scoring uses the frozen judgments finalized after human review of disagreements between Astra and Kimi-K3. A separate sensitivity analysis evaluates the hypothetical adoption of all second-model objections and decisions on formerly unresolved candidates.
Each of 2,366 deduplicated candidates from 939 human-data predictions is separately submitted to Kimi-K3 with the original prediction, gold, relevant events and reference turns, and Astra's judgment and reason. Kimi must identify a specific first-round error, applicable rule, and locatable evidence before contradicting it. Another permissible interpretation or naming preference is not an objection. For formerly unresolved candidates, it returns a binary decision when supported, otherwise identifies missing context. Because it sees the first opinion, this is sequential verification, not blind independent agreement.

\paragraph{Frozen scoring and scope.}
Records retain candidate IDs, both judgments, rules, and evidence positions. Disagreements between the two models are resolved through human review before the judgments used for primary scoring are frozen. Confirmed equivalent expressions participate normally in required-event matching, not as exempt extras. Mappings apply only to their sample and candidate; they are not global aliases and cannot change speaker, action, time, cycle, or source-edge constraints. Deterministic code uses the finalized expression judgments for structure checks, maximum one-to-one matching, and aggregation.

\paragraph{Coverage and sensitivity.}
All 2,366 candidates received second-stage responses: 2,108 did not meet the clear-error objection threshold, 154 received contrary judgments, 103 formerly unresolved candidates received decisions, and one remained unresolved at the second-model stage. Contrary judgments comprise 129 non-equivalent-to-equivalent and 25 equivalent-to-non-equivalent decisions. These counts summarize judgments at the second-model review stage.

The all-adopted sensitivity condition applies the 154 objections and 103 decisions while freezing gold, predictions, structure rules, and extra-node judgments. It measures the effect of the second model's proposals, not confirmed first-round error. Across 41 conditions, the maximum absolute DCIST change is 1.81 pp. Astra's unlabeled-script score rises from 28.99\% to 30.80\%, with seven trajectories becoming correct (100 to 107 matches). Its labeled-script score rises from 38.72\% to 40.50\%: ten new matches and two lost matches, 12 changed outcomes out of 308 (3.90\%), and 133 to 141 total matches. These changes indicate relative stability within this sequential review while leaving the full-recovery challenge intact.

\paragraph{Human review and traceability.}
Candidates on which Astra and Kimi-K3 disagree are reviewed by a human, who makes the final equivalent or non-equivalent judgment. A single human reviewer also independently examined every trajectory whose match outcome changed. Prediction and gold coordinates, both model judgments, and score-change records are retained for each case to support traceable inspection.

\subsection{Additional-Content Adjudication}
After expression matching, code identifies remaining unmatched nodes and absent-gold fields. Astra reviews them against actual model input, context, reference script, and frozen task rules. Supported, permitted refinements are exempt. Reviewed candidates with a specific unresolved evidence or rule boundary are also exempt; unreviewed candidates are not. Repetition without change, ordinary questions, cross-field reuse of a conclusion, and unsupported or rule-violating insertions are not exempt. A mentioned number alone does not justify an event: action, field, value, speaker, scope, time, and relations all matter. Any erroneous insertion invalidates its trajectory.

For a gold-corresponding trajectory, exempt nodes are removed only from the scoring view and predecessor paths are contracted. This cannot fix unsupported edges, cycle misalignment, substituted required events, or missing required events. For a wholly new independent field, all predicted nodes and relations are reviewed without inventing unannotated later events. If all output content is valid, or reviewed evidence remains unresolved without a definite error, the new trajectory is exempt from the denominator but not counted as a true positive. Any invalid node or relation prevents whole-trajectory exemption. Renaming an existing gold event to a related field is not new independent information.

Whole-trajectory records and expanded node records are deduplicated by original coordinates; rejecting a whole trajectory does not classify every constituent node as wrong. Actual cascade input takes precedence for what the model observed. A coarse script cannot resolve a discrepancy requiring listening and cannot repair the transcript or answer. Such evidence remains unresolved. Prediction and adjudication sessions are separate, but an Astra prediction and its Astra review still use the same model. The protocol is model-assisted semantic/evidence adjudication followed by deterministic scoring, not entirely mechanical scoring.

\subsection{Aggregation and Verification}
Predicted evidence uses $|P\cap G|/|P|\geq0.80$, with positive predicted duration. Every required event must pass. After exemptions and edge contraction, maximum one-to-one trajectory matching supplies $TP_i$ for the main equations. Global micro-F1 is supplementary; actions have no extra weights. Complete malformed answers are output errors. Cohort-specific completion accounting follows Appendix~\ref{app:gemini} for direct audio and the reported completed-output cohorts for the earlier experiments.

Let $U$ contain all parseable predicted nodes and $E$ the union of individually exempt nodes and nodes in wholly exempt new trajectories. The auxiliary ratio is
\begin{equation}R_{\mathrm{extra}}=|E|/|U|.\end{equation}
Unique coordinates include input condition, recording, trajectory, cycle, and event position. A three-node new trajectory contributes three nodes, not one. Legitimate extras can remain in $E$ even if another error invalidates their trajectory. This ratio is neither a score bonus nor normalized by gold-node count. The human-data extra-node ratio uses completed adjudication coverage. Synthetic review closes all score-relevant candidates through explicit judgments or verified score-invariance checks; it does not provide an exhaustive extra-node classification.

Frozen reproduction tests cover money equivalence, the 80\% time rule, identity/actions, edges, cycles, exemptions, and one-to-one matching. The manuscript's scoring snapshot passed 40 regression tests and self-matching for all 26 gold files with and without time. Sequential-review sensitivity reproduction recovered all 939 baseline results, checked matching counts and F1 arithmetic, and verified frozen-input and judgment hashes. Review records preserve original positions, evidence, rule references, and reasons.

\section{Statistical Methods and Full Results}\label{app:statistics}
\subsection{Clustered Intervals and Paired Tests}
There are 16 human dialogue families: each A/B pair is one cluster, and each independent long script is one. For a comparison with $K$ families, resample $K$ families with replacement, include all available recordings of each selected family, and recompute recording-macro DCIST. We use 20,000 draws and percentile 2.5/97.5 limits. This preserves recording weights instead of first averaging families equally. The main 25-recording cohort still contains 16 families; paired A/B diagnostics require both variants.

The main-system tests use the fixed 25-recording cohort and its completion accounting in Appendix~\ref{app:gemini}; earlier paired ablations retain their jointly completed cohorts. Differences for every recording within a family receive the same sign, with all $2^K$ assignments enumerated. Two-sided $p$ is the fraction with an absolute recording-macro difference at least the observed value. All main comparisons contain 16 families and enumerate 65,536 assignments. Holm correction jointly covers the 21 contrasts between seven main systems. Four label contrasts, two B2 contrasts, three historical A contrasts, and three attribution controls remain separate test families. Intervals are pointwise, not simultaneous; significance uses corrected $p$.

For sorted raw values $p_{(1)}\leq\cdots\leq p_{(21)}$, the adjusted value at rank $j$ is $\min\{1,\max_{k\leq j}(22-k)p_{(k)}\}$. Correction uses unrounded values. Before extending the family, all ten original cascade raw and adjusted $p$ values were exactly reproduced. All 21 new tail counts were verified by an independent enumeration. The two original significant cascade contrasts remain significant under the larger family; three contrasts involving direct-audio models also pass 0.05 (Table~\ref{tab:system_tests}).

The base seed is 20261013; name hashes derive reproducible independent streams for statistical quantities. All predictions, gold, mappings, and extra judgments stay fixed. Differences are computed before rounding; subtracting displayed scores can therefore differ by 0.01 pp. Intervals reflect sampled dialogue-family variation, not repeated generation or independent human-participant sampling.

\input{results_system_tests.tex}
\input{results_cascade_intervals.tex}
\input{results_script_intervals.tex}
\input{results_noise_tests.tex}
Historical FunASR A1 declines by 4.50 pp with corrected $p=0.047$; FunASR A2 and VibeVoice A1 do not meet 0.05. None of the fixed automatic-speaker controls meets that threshold. Therefore the numerical speaker-control changes alone do not establish a segmentation effect.

\subsection{Terminal-Value Projection}
The final required event in the final cycle supplies each non-null gold terminal value. There are 280 such fields across all 26 recordings and 266 in the main cohort; 28 null-valued fields are excluded from the all-recording diagnostic. Field and value comparisons retain exact labels, frozen sample-local mappings, and monetary conversion; no new equivalences are adjudicated for this projection.

Predictions retain their original event order. Required matches remain, exempt refinements are removed in the scoring view, and the remaining terminal value is projected. Field assignment is one-to-one, prioritizing existing full-chain matches; one prediction cannot hit multiple gold fields. The output-format gate remains, but the diagnostic does not additionally require intermediate events, action, speaker, scope, or time. It is a projection of the same answer, not a separately prompted model task.

For eligible gold fields $D$, terminal-value hits $V$, and full-chain hits $T$, we report $|V|/|D|$, $|T|/|D|$, and $|V\setminus T|/|V|$. All observations satisfy $T\subseteq V$. Frozen records retain denominators, field coordinates, projected prediction nodes, and mapping IDs.
\input{results_terminal.tex}
\input{results_terminal_intervals.tex}

\subsection{A/B Joint Recovery}
Pair variants by family metadata and match shared entity--field keys. For shared field pairs $Q$,
\begin{equation}
J_{AB}=\frac{1}{|Q|}\sum_{q\in Q}\mathbf{1}[q_A\text{ matches}\land q_B\text{ matches}].
\end{equation}
Each variant must satisfy its own full gold trajectory; the two gold records need not be identical. Fields unique to one variant remain in DCIST but not this diagnostic. One audio family lacks a completed variant, leaving nine pairs/87 shared fields; scripts retain ten/102.

The changed subset compares cycles, actions, speakers, scopes, values, settlement results, and predecessor positions, normalizing explicit monetary units and ignoring renamable IDs, reasons, and evidence times. This yields 35 audio and 40 script field pairs. It adds no semantic judgment. Intervals resample paired families and aggregate field-pair counts.
\input{results_ab.tex}

\section{Stratification and Local Diagnostics}\label{app:additional}
\subsection{Speakers, Duration, and Chain Length}
Speaker-count strata contain five two-speaker, thirteen three-speaker, and seven four-speaker recordings. Duration strata contain nine recordings below 15 minutes, twelve from 15 to below 30, and four of at least 30 minutes. Each dimension partitions the same 25-recording cohort and macro-averages F1 within strata. Scenario, length, and speaker count are not independently controlled; four long recordings cannot establish a stable length-degradation law.
\input{results_strata.tex}
\input{results_chain_length.tex}
Chain length counts required gold events across all cycles, not predicted refinements. Recall is complete matches divided by gold trajectories in that bin; unmatched predicted fields are not arbitrarily allocated to a length bin. One-node recall spans 7.63--42.75\%, while three-or-more-node recall spans 1.69--13.56\% across systems.

\subsection{Extra Nodes and Micro-Averaging}
\input{results_auxiliary.tex}
The exemption union deduplicates node coordinates and includes supported or reviewed-unresolved extras. Nodes selected as required matches are removed from that union; a formerly new trajectory matched to gold is no longer wholly exempt. All nodes in parseable output form the denominator, even if that output fails a schema check. Unparseable output contributes no invented node count but is still scored by the format rule. Micro-F1 is $2\sum_iTP_i/(\sum_iG_i+\sum_i(P_i-L_i))$, weighting recordings with more trajectories more heavily than recording-macro DCIST. The extra proportion describes granularity, not hallucination, and cannot cancel omissions.

\begin{table*}[t]
\centering\small
\begin{tabularx}{\textwidth}{@{}p{.20\textwidth}XXX@{}}
\toprule
Field / evidence & Original input and prediction & B2 input and prediction & Located mismatch \\
\midrule
Financing amount; R0023; 674.97--680.23 s & ``Set financing at 8.4 million''; 9.0M start, 8.4M settlement & ``Set financing at 2.4 million''; 9.0M start, 2.4M settlement & Wrong settlement value; other required attributes, edges, and time pass \\
Service fee; R0023; 713.93--720.30 s & ``Set the fee rate at 0.8\%''; 1\% start, 0.8\% settlement & ``Set the fee rate at 1.8\%''; 1\% start, 1.8\% settlement & Wrong settlement value despite matching remaining requirements \\
Requested amount and term; R0013; 263.43--267.55 s & ``Five million, six months''; prediction contains both & Corrupted phrase corresponding to ``five million user apply for him''; prediction retains 5M, omits six months & Missing term in a composite value; correct amount alone is insufficient \\
\bottomrule
\end{tabularx}
\caption{Traceable Kimi-K3 examples, translated into English for presentation. The first two use VibeVoice, the third FunASR. Times are intervals cited by the original predictions. Each original trajectory matches and its B2 counterpart fails. Actual input errors are preserved rather than replaced with scripted text. M denotes million CNY.}\label{tab:critical_examples}
\end{table*}

\subsection{Time Windows and Speaker Identifiers}
\input{results_protocol_sensitivity.tex}
\input{results_time_windows.tex}
Each diagnostic changes one mechanical requirement while holding predictions, gold, semantic mappings, and extra judgments fixed. No previously unmapped expression becomes equivalent automatically. Moving alignment anchors introduced no blocked, newly unreviewed extra candidates in these time diagnostics. Since mappings were established under the 80\% protocol, the no-time diagnostic is not a newly adjudicated upper bound.

Valid predicted windows have median lengths of 5.06--10.20 seconds. Of 2,673 windows, only seven are at most one second. Within fully matched trajectories, median predicted/gold window-length ratios range from 15.54\% to 48.56\%. Requiring coverage of at least half the gold window, in addition to the official 80\% predicted-window criterion, reduces scores to 0.00--3.99\%. Coarse evidence windows often exceed the cited phrase; this added constraint measures boundary coverage rather than simply correct contextual location.

The identifier diagnostic enumerates one globally consistent, one-to-one mapping per recording; it neither merges speakers nor changes them independently per event. It enumerates only when the numbers of distinct gold/predicted event speakers agree and are at most six, otherwise retaining original IDs. All other constraints remain. One FunASR+Astra recording and four FunASR+Kimi recordings improve; the other three systems have no improvement. This diagnoses numbering convention, not diarization accuracy.

\subsection{Local Evidence and Trajectory Changes}
\input{results_noise_localization.tex}
The same-route Kimi comparison joins conditions by recording and gold-trajectory coordinate, not prediction order. It covers 26 FunASR recordings/308 trajectories and 22 VibeVoice recordings/255 trajectories. We retain lost and gained matches as well as stable outcomes, rather than selecting only degradations.

Local diagnostics reuse frozen B2 anchors linked by event ID and require the anchor core to overlap the revised gold window. An absence signal means a normalized anchor form occurs in original local context but not B2 context; normalization covers explicit numeric spelling, punctuation, and unit forms. FunASR provides native word times, whereas VibeVoice uses overlapping segment text. All 26 FunASR lost trajectories and 13 of 16 VibeVoice lost trajectories contain such a signal. It also occurs in eleven and one gained trajectories, respectively. It is therefore neither a sufficient cause nor a manually verified ASR error rate.

The wrong amount and fee appear directly in both noisy input and settlement predictions; the third case omits a term within a composite value. Other context, segmentation, and downstream inference can change concurrently, and both recovered and lost matches occur. These examples locate specific observed errors, not a claim that every score difference is caused by one word.

\begin{table*}[!t]
\input{astra_script_errors.tex}
\end{table*}

\section{Astra Error Distributions on Reference Scripts}\label{app:astra_script_errors}
Table~\ref{tab:astra_script_errors} reports Astra's error distributions under the two reference-script conditions. The analysis covers all originally unmatched gold trajectories: 175 with speaker labels and 208 without speaker labels, out of 308 gold trajectories across 26 scripts in each condition. Counts refer to trajectories involving each error type, rather than the number of erroneous nodes.

\noindent Content and scope errors are the two most frequent categories in both conditions. Speaker errors affect 38 unmatched trajectories (21.71\%) with labels and 72 (34.62\%) without labels. Missing required nodes and erroneous extra nodes also occur in both conditions, whereas a cycle-count mismatch appears in only one labeled-script trajectory. Since these categories can co-occur and the two failure sets differ, the percentages describe the composition of each set, not mutually exclusive causes or a paired causal decomposition of the score difference.

%% file: results_synthetic_all.tex
\begin{table*}[t]
\centering\small
\begin{tabular}{@{}llrrrr@{}}
\toprule
ASR & Reasoner & Completed / total & Format errors & DCIST & No time \\
\midrule
FunASR & DeepSeek & 2232/2246 & 66 & 32.65 & 51.88 \\
FunASR & Qwen & 2245/2246 & 95 & 28.17 & 46.25 \\
VibeVoice & DeepSeek & 2239/2246 & 53 & 22.60 & 74.44 \\
VibeVoice & Qwen & 2241/2246 & 65 & 21.66 & 67.57 \\
\bottomrule
\end{tabular}
\caption{Synthetic results over all completed outputs per system, with coverage reported separately. DeepSeek denotes DeepSeek V4 Flash and Qwen denotes Qwen3.8-Max. Complete malformed answers remain scored; incomplete outputs are not imputed as zeros.}\label{tab:synthetic_all}
\end{table*}

%% file: results_synthetic_decomposition.tex
\begin{table*}[t]
\centering\small
\begin{tabular}{@{}lllrrrr@{}}
\toprule
ASR & Reasoner & Time & Legacy & + Amount & + Semantics & + Extras \\
\midrule
FunASR & DeepSeek & Yes & 15.42 & 15.62 & 31.90 & 32.71 \\
FunASR & DeepSeek & No & 24.16 & 24.45 & 50.55 & 51.98 \\
FunASR & Qwen & Yes & 6.47 & 6.75 & 28.10 & 28.33 \\
FunASR & Qwen & No & 10.38 & 10.92 & 46.08 & 46.47 \\
VibeVoice & DeepSeek & Yes & 10.95 & 10.97 & 22.18 & 22.61 \\
VibeVoice & DeepSeek & No & 36.18 & 36.21 & 72.96 & 74.59 \\
VibeVoice & Qwen & Yes & 5.97 & 6.00 & 21.59 & 21.64 \\
VibeVoice & Qwen & No & 17.65 & 17.80 & 67.41 & 67.70 \\
\bottomrule
\end{tabular}
\caption{Sequential scoring decomposition on the same 2,224 recordings, keeping predictions and gold fixed. Columns cumulatively introduce monetary equivalence, reviewed expression mappings, and extra-node adjudication; the last column is the reported score. Increments depend on this order and are not independent causal effects.}\label{tab:synthetic_decomposition}
\end{table*}

%% file: results_noise_history.tex
\begin{table*}[t]
\centering
\fontsize{10}{12}\selectfont
\begin{tabular}{@{}lrrrrrr@{}}
\toprule
Frontend & Condition & N & CER (\%) & WER (\%) & cpWER (\%) & DCIST (\%) \\
\midrule
FunASR & Original & 26 & 17.12 & 21.69 & 31.18 & 12.41 \\
FunASR & A1 & 26 & 19.20 & 24.17 & 34.59 & 7.91 \\
FunASR & A2 & 26 & 23.67 & 29.51 & 43.00 & 10.19 \\
VibeVoice & Original & 23 & 9.10 & 11.13 & 13.33 & 10.10 \\
VibeVoice & A1 & 23 & 10.09 & 12.34 & 14.23 & 11.22 \\
\bottomrule
\end{tabular}
\caption{Historical Kimi-K3 evidence-external noise experiments: 26 common FunASR recordings and 23 common VibeVoice recordings. Transcription metrics aggregate corpus-level errors against reading scripts. These routes differ from the B2 control.}\label{tab:noisehistory}
\end{table*}

%% file: results_speaker_noise.tex
\begin{table*}[t]
\centering
\fontsize{10}{12}\selectfont
\begin{tabular}{@{}lrrrr@{}}
\toprule
Kimi-K3 condition & N & Automatic (\%) & Fixed (\%) & Difference \\
\midrule
FunASR A1 & 26 & 7.91 & 9.73 & +1.82 \\
FunASR A2 & 26 & 10.19 & 8.94 & -1.25 \\
VibeVoice A1 & 23 & 11.22 & 12.46 & +1.24 \\
\bottomrule
\end{tabular}
\caption{Automatic-speaker control. Fixed attribution uses speaker predictions from the unperturbed recording, not human gold identities. Differences are percentage points.}\label{tab:speakercontrol}
\end{table*}

%% file: results_system_tests.tex
\begin{table*}[t]
\centering\small
\begin{tabular}{@{}lrrr@{}}
\toprule
Comparison (first minus second) & Difference (pp) & Raw $p$ & $p_{\rm Holm}$ \\
\midrule
F+Astra $-$ F+Kimi & +11.55 & 0.0029 & \textbf{0.0498} \\
F+Astra $-$ F+Qwen & +15.59 & 0.0020 & \textbf{0.0371} \\
F+Astra $-$ V+Kimi & +10.26 & 0.0078 & 0.1172 \\
F+Astra $-$ V+Qwen & +15.62 & 0.0137 & 0.1777 \\
F+Astra $-$ G-Argon & +2.53 & 0.0703 & 0.7031 \\
F+Astra $-$ G-Flash & +14.75 & 0.0010 & \textbf{0.0205} \\
F+Kimi $-$ F+Qwen & +4.04 & 0.3906 & 1.0000 \\
F+Kimi $-$ V+Kimi & -1.29 & 0.5049 & 1.0000 \\
F+Kimi $-$ V+Qwen & +4.07 & 0.6211 & 1.0000 \\
F+Kimi $-$ G-Argon & -9.02 & 0.0068 & 0.1094 \\
F+Kimi $-$ G-Flash & +3.20 & 0.5234 & 1.0000 \\
F+Qwen $-$ V+Kimi & -5.33 & 0.0547 & 0.6016 \\
F+Qwen $-$ V+Qwen & +0.03 & 0.9961 & 1.0000 \\
F+Qwen $-$ G-Argon & -13.07 & 0.0010 & \textbf{0.0205} \\
F+Qwen $-$ G-Flash & -0.84 & 0.5312 & 1.0000 \\
V+Kimi $-$ V+Qwen & +5.36 & 0.2266 & 1.0000 \\
V+Kimi $-$ G-Argon & -7.73 & 0.0088 & 0.1230 \\
V+Kimi $-$ G-Flash & +4.49 & 0.1641 & 1.0000 \\
V+Qwen $-$ G-Argon & -13.09 & 0.0195 & 0.2344 \\
V+Qwen $-$ G-Flash & -0.87 & 0.7070 & 1.0000 \\
G-Argon $-$ G-Flash & +12.23 & 0.0020 & \textbf{0.0371} \\
\bottomrule
\end{tabular}
\caption{All 21 pairwise main-system comparisons on the fixed 25-recording cohort in 16 families. F/V denote FunASR/VibeVoice; G-Argon/G-Flash denote Gemini-4-Argon/Gemini-3.8-Flash. Two-sided exact family-level sign-permutation tests enumerate 65,536 assignments. Holm correction covers all 21 contrasts jointly; bold adjusted values pass 0.05 before rounding. Differences are recording-macro DCIST percentage points.}\label{tab:system_tests}
\end{table*}

%% file: results_cascade_intervals.tex
\begin{table*}[t]
\centering\small
\begin{tabular}{@{}lrr@{}}
\toprule
System / contrast & Estimate (\% / pp) & Pointwise 95\% CI \\
\midrule
F+Astra & 21.35 & [9.70, 34.62] \\
F+Kimi & 9.80 & [3.08, 19.52] \\
F+Qwen & 5.75 & [1.16, 11.19] \\
V+Kimi & 11.09 & [4.46, 19.17] \\
V+Qwen & 5.73 & [2.30, 9.46] \\
G-Argon & 18.82 & [8.55, 30.71] \\
G-Flash & 6.59 & [1.60, 12.77] \\
\midrule
F+Astra $-$ F+Kimi & +11.55 & [5.12, 18.69] \\
F+Astra $-$ F+Qwen & +15.59 & [6.84, 25.79] \\
F+Astra $-$ V+Kimi & +10.26 & [3.98, 17.46] \\
F+Astra $-$ V+Qwen & +15.62 & [4.60, 29.29] \\
F+Astra $-$ G-Argon & +2.53 & [0.32, 4.81] \\
F+Astra $-$ G-Flash & +14.75 & [5.80, 25.49] \\
F+Kimi $-$ F+Qwen & +4.04 & [-2.91, 11.87] \\
F+Kimi $-$ V+Kimi & -1.29 & [-4.99, 2.19] \\
F+Kimi $-$ V+Qwen & +4.07 & [-3.45, 15.04] \\
F+Kimi $-$ G-Argon & -9.02 & [-15.47, -3.64] \\
F+Kimi $-$ G-Flash & +3.20 & [-4.84, 12.18] \\
F+Qwen $-$ V+Kimi & -5.33 & [-11.24, -0.60] \\
F+Qwen $-$ V+Qwen & +0.03 & [-4.00, 4.36] \\
F+Qwen $-$ G-Argon & -13.07 & [-22.00, -5.56] \\
F+Qwen $-$ G-Flash & -0.84 & [-3.28, 1.37] \\
V+Kimi $-$ V+Qwen & +5.36 & [-1.15, 14.45] \\
V+Kimi $-$ G-Argon & -7.73 & [-13.60, -2.77] \\
V+Kimi $-$ G-Flash & +4.49 & [-0.74, 10.88] \\
V+Qwen $-$ G-Argon & -13.09 & [-25.28, -3.51] \\
V+Qwen $-$ G-Flash & -0.87 & [-5.38, 3.30] \\
G-Argon $-$ G-Flash & +12.23 & [4.70, 21.40] \\
\bottomrule
\end{tabular}
\caption{Family-clustered 95\% confidence intervals for all seven main systems and their 21 pairwise differences on the fixed 25-recording cohort (16 families; 20,000 bootstrap draws). F/V denote FunASR/VibeVoice; G-Argon/G-Flash denote Gemini-4-Argon/Gemini-3.8-Flash. System estimates are percentages; differences and their intervals are percentage points. These are pointwise, not simultaneous, intervals. Main-system significance is determined by the 21-contrast Holm correction in Table~\ref{tab:system_tests}, not by interval exclusion of zero.}\label{tab:cascade_intervals}
\end{table*}

%% file: results_script_intervals.tex
\begin{table*}[t]
\centering
\fontsize{9.5}{11.5}\selectfont
\begin{tabular}{@{}lrrrr@{}}
\toprule
Model & Labeled 95\% CI & Unlabeled 95\% CI & Difference 95\% CI & $p_{\rm Holm}$ \\
\midrule
Astra & [26.06, 52.53] & [16.77, 42.44] & [4.18, 15.14] & 0.019 \\
Kimi-K3 & [13.44, 34.36] & [16.21, 31.47] & [-5.90, 4.24] & 0.766 \\
Qwen3.8-Max & [16.36, 21.23] & [7.42, 20.30] & [-1.11, 11.20] & 0.285 \\
DeepSeek V4 Flash & [9.31, 19.02] & [3.25, 11.60] & [3.05, 10.56] & 0.019 \\
\bottomrule
\end{tabular}
\caption{Family-clustered intervals and two-sided paired tests for 26 scripts in 16 families. Differences are labeled minus unlabeled, in percentage points. Holm correction jointly covers the four within-model label contrasts.}\label{tab:script_intervals}
\end{table*}

%% file: results_noise_tests.tex
\begin{table*}[t]
\centering
\fontsize{9.5}{11.5}\selectfont
\begin{tabular}{@{}lrrrrr@{}}
\toprule
Kimi-K3 comparison & N & Families & Difference & 95\% CI & $p_{\rm Holm}$ \\
\midrule
FunASR: B2 $-$ Original & 26 & 16 & -2.04 & [-8.15, 2.40] & 0.625 \\
VibeVoice: B2 $-$ Original & 22 & 16 & -3.32 & [-5.51, -0.96] & 0.039 \\
FunASR: A1 $-$ Original & 26 & 16 & -4.50 & [-9.08, -0.81] & 0.047 \\
FunASR: A2 $-$ Original & 26 & 16 & -2.22 & [-6.05, 0.85] & 0.562 \\
VibeVoice: A1 $-$ Original & 23 & 15 & +1.13 & [-1.71, 4.72] & 0.609 \\
FunASR A1: Fixed $-$ Automatic & 26 & 16 & +1.82 & [-0.11, 4.32] & 0.469 \\
FunASR A2: Fixed $-$ Automatic & 26 & 16 & -1.25 & [-3.34, 0.00] & 0.625 \\
VibeVoice A1: Fixed $-$ Automatic & 23 & 15 & +1.24 & [-0.48, 3.31] & 0.625 \\
\bottomrule
\end{tabular}
\caption{Paired acoustic and automatic-speaker controls. Holm correction is separate for two B2 comparisons, three historical A comparisons, and three speaker controls. Only jointly completed recordings enter each contrast. Differences and intervals are percentage points. Fixed speaker labels are still automatic.}\label{tab:noise_tests}
\end{table*}

%% file: results_terminal.tex
\begin{table*}[t]
\centering
\fontsize{9.5}{11.5}\selectfont
\begin{tabular}{@{}lrrrr@{}}
\toprule
System / script input & Fields & Value recall (\%) & Chain recall (\%) & Value-only hits (\%) \\
\midrule
FunASR + Astra & 266 & 62.41 & 30.45 & 51.20 \\
FunASR + Kimi-K3 & 266 & 52.63 & 14.29 & 72.86 \\
FunASR + Qwen3.8-Max & 266 & 32.71 & 6.77 & 79.31 \\
VibeVoice + Kimi-K3 & 266 & 51.50 & 14.66 & 71.53 \\
VibeVoice + Qwen3.8-Max & 266 & 35.71 & 5.26 & 85.26 \\
Astra (labeled) & 280 & 68.93 & 46.43 & 32.64 \\
Kimi-K3 (labeled) & 280 & 51.07 & 27.86 & 45.45 \\
Qwen3.8-Max (labeled) & 280 & 51.79 & 19.29 & 62.76 \\
DeepSeek V4 Flash (labeled) & 280 & 45.00 & 14.64 & 67.46 \\
Astra (unlabeled) & 280 & 66.07 & 34.64 & 47.57 \\
Kimi-K3 (unlabeled) & 280 & 58.57 & 27.14 & 53.66 \\
Qwen3.8-Max (unlabeled) & 280 & 42.86 & 14.29 & 66.67 \\
DeepSeek V4 Flash (unlabeled) & 280 & 28.57 & 6.43 & 77.50 \\
\bottomrule
\end{tabular}
\caption{Terminal-value projections versus full trajectories from the same predictions. Both recall denominators are gold fields with non-null terminal values: 266 fields for the common 25 audio recordings, 280 for all 26 scripts. The last column divides value-correct but chain-incorrect fields by all value-correct fields. Full-chain recall is not macro-F1 DCIST.}\label{tab:terminal}
\end{table*}

%% file: results_terminal_intervals.tex
\begin{table*}[t]
\centering
\fontsize{9.5}{11.5}\selectfont
\begin{tabular}{@{}lrr@{}}
\toprule
System / script input & Terminal-value recall 95\% CI & Full-chain recall 95\% CI \\
\midrule
FunASR + Astra & [47.77, 74.80] & [13.08, 45.08] \\
FunASR + Kimi-K3 & [37.96, 64.91] & [4.09, 25.82] \\
FunASR + Qwen3.8-Max & [20.93, 44.24] & [1.90, 11.30] \\
VibeVoice + Kimi-K3 & [32.14, 66.99] & [5.29, 25.25] \\
VibeVoice + Qwen3.8-Max & [25.16, 45.87] & [2.17, 8.66] \\
Astra (labeled) & [55.00, 79.80] & [30.70, 60.36] \\
Kimi-K3 (labeled) & [31.17, 67.59] & [13.66, 43.20] \\
Qwen3.8-Max (labeled) & [43.03, 60.11] & [15.84, 23.20] \\
DeepSeek V4 Flash (labeled) & [27.82, 60.12] & [9.09, 19.93] \\
Astra (unlabeled) & [54.55, 75.30] & [20.92, 49.50] \\
Kimi-K3 (unlabeled) & [45.00, 69.37] & [16.05, 38.46] \\
Qwen3.8-Max (unlabeled) & [31.06, 52.90] & [8.11, 22.28] \\
DeepSeek V4 Flash (unlabeled) & [17.36, 40.45] & [2.72, 11.71] \\
\bottomrule
\end{tabular}
\caption{Family-clustered 95\% confidence intervals for terminal-value and full-chain recall on the same non-null-terminal-value subsets as Table~\ref{tab:terminal}.}\label{tab:terminal_intervals}
\end{table*}

%% file: results_ab.tex
\begin{table*}[t]
\centering
\fontsize{9.5}{11.5}\selectfont
\begin{tabular}{@{}lrrrrr@{}}
\toprule
System / script input & Field pairs & Both correct & Joint (\%) & 95\% CI & Changed subset \\
\midrule
FunASR + Astra & 87 & 21 & 24.14 & [0.00, 41.18] & 3/35 \\
FunASR + Kimi-K3 & 87 & 13 & 14.94 & [0.00, 29.47] & 0/35 \\
FunASR + Qwen3.8-Max & 87 & 3 & 3.45 & [0.00, 7.81] & 1/35 \\
VibeVoice + Kimi-K3 & 87 & 5 & 5.75 & [0.00, 13.13] & 0/35 \\
VibeVoice + Qwen3.8-Max & 87 & 1 & 1.15 & [0.00, 4.00] & 0/35 \\
Astra (labeled) & 102 & 41 & 40.20 & [13.89, 61.07] & 14/40 \\
Kimi-K3 (labeled) & 102 & 23 & 22.55 & [7.87, 37.17] & 2/40 \\
Qwen3.8-Max (labeled) & 102 & 12 & 11.76 & [6.15, 17.58] & 0/40 \\
DeepSeek V4 Flash (labeled) & 102 & 14 & 13.73 & [7.14, 19.61] & 2/40 \\
Astra (unlabeled) & 102 & 26 & 25.49 & [6.15, 43.59] & 4/40 \\
Kimi-K3 (unlabeled) & 102 & 20 & 19.61 & [8.14, 31.30] & 1/40 \\
Qwen3.8-Max (unlabeled) & 102 & 12 & 11.76 & [5.48, 18.31] & 0/40 \\
DeepSeek V4 Flash (unlabeled) & 102 & 6 & 5.88 & [1.31, 13.25] & 0/40 \\
\bottomrule
\end{tabular}
\caption{A/B joint recovery. Audio systems share nine complete families and 87 field pairs; scripts cover ten families and 102 pairs. A pair succeeds only if both trajectories match. The changed subset has differing gold structure or attributes after ignoring evidence times and renamable IDs (35 and 40 pairs). Intervals resample paired families; fields unique to one variant are excluded.}\label{tab:ab}
\end{table*}

%% file: results_strata.tex
\begin{table*}[!t]
\centering
\fontsize{9.5}{11.5}\selectfont
\begin{tabular}{@{}lrrrrrrr@{}}
\toprule
Stratum & Audio & Tracks & F+A & F+K & F+Q & V+K & V+Q \\
\midrule
2 speakers & 5 & 40 & 20.00 & 6.98 & 12.00 & 12.68 & 10.68 \\
3 speakers & 13 & 163 & 20.24 & 5.68 & 3.64 & 8.77 & 4.99 \\
4 speakers & 7 & 90 & 24.36 & 19.46 & 5.21 & 14.25 & 3.57 \\
$<15$ min & 9 & 68 & 5.84 & 1.39 & 2.22 & 3.02 & 5.25 \\
15--$<30$ min & 12 & 163 & 34.88 & 18.21 & 9.25 & 19.25 & 6.98 \\
$\geq30$ min & 4 & 62 & 15.62 & 3.49 & 3.23 & 4.76 & 3.04 \\
\bottomrule
\end{tabular}
\caption{DCIST (\%) by speaker count and duration on the same 25 recordings. F/V denote FunASR/VibeVoice; A/K/Q denote Astra/Kimi-K3/Qwen3.8-Max. Each stratum macro-averages recording F1. Counts refer to script roles and full audio duration; the middle duration bin excludes 30 minutes.}\label{tab:strata}
\end{table*}

%% file: results_chain_length.tex
\begin{table*}[!t]
\centering
\fontsize{9.5}{11.5}\selectfont
\begin{tabular}{@{}lrrrrrr@{}}
\toprule
Nodes & Tracks & F+A & F+K & F+Q & V+K & V+Q \\
\midrule
1 & 131 & 56/131 (42.75) & 31/131 (23.66) & 15/131 (11.45) & 29/131 (22.14) & 10/131 (7.63) \\
2 & 103 & 18/103 (17.48) & 6/103 (5.83) & 1/103 (0.97) & 9/103 (8.74) & 2/103 (1.94) \\
3+ & 59 & 8/59 (13.56) & 1/59 (1.69) & 2/59 (3.39) & 1/59 (1.69) & 3/59 (5.08) \\
\bottomrule
\end{tabular}
\caption{Full-chain recall by required gold-node count: hits/trajectories (\%). F/V denote FunASR/VibeVoice; A/K/Q denote Astra/Kimi-K3/Qwen3.8-Max. These are gold-trajectory recall rates, not stratified DCIST.}\label{tab:chain_length}
\end{table*}

%% file: results_auxiliary.tex
\begin{table*}[!t]
\centering
\fontsize{9.5}{11.5}\selectfont
\begin{tabular}{@{}lrrrrrr@{}}
\toprule
System / script input & N & Macro F1 & Micro F1 & All nodes & Exempt nodes & Extra (\%) \\
\midrule
FunASR + Astra & 25 & 21.35 & 27.06 & 732 & 193 & 26.37 \\
FunASR + Kimi-K3 & 25 & 9.80 & 13.36 & 483 & 73 & 15.11 \\
FunASR + Qwen & 25 & 5.75 & 6.36 & 476 & 50 & 10.50 \\
VibeVoice + Kimi-K3 & 25 & 11.09 & 13.66 & 493 & 76 & 15.42 \\
VibeVoice + Qwen & 25 & 5.73 & 5.23 & 491 & 66 & 13.44 \\
Astra / labeled & 26 & 38.72 & 43.18 & 744 & 203 & 27.28 \\
Kimi-K3 / labeled & 26 & 22.72 & 26.71 & 495 & 89 & 17.98 \\
Qwen / labeled & 26 & 18.89 & 18.77 & 524 & 114 & 21.76 \\
DeepSeek / labeled & 26 & 14.04 & 13.99 & 512 & 90 & 17.58 \\
Astra / unlabeled & 26 & 28.99 & 32.21 & 755 & 198 & 26.23 \\
Kimi-K3 / unlabeled & 26 & 23.58 & 25.75 & 554 & 108 & 19.49 \\
Qwen / unlabeled & 26 & 13.79 & 13.05 & 531 & 101 & 19.02 \\
DeepSeek / unlabeled & 26 & 7.31 & 6.01 & 478 & 85 & 17.78 \\
\bottomrule
\end{tabular}
\caption{Auxiliary metrics for the common audio cohort (25 recordings) and scripts (26 per condition). Macro F1 is DCIST. F1 and extra-node proportions are percentages. Node denominators include parseable outputs; no nodes are invented for unparseable answers, which remain subject to the format-error rule.}\label{tab:auxiliary}
\end{table*}

%% file: results_protocol_sensitivity.tex
\begin{table*}[!t]
\centering
\fontsize{9.5}{11.5}\selectfont
\begin{tabular}{@{}lrrrrrr@{}}
\toprule
System & 50\% & 80\% & 90\% & No time & Bidirectional & ID permutation \\
\midrule
FunASR + Astra & 21.74 & 21.35 & 20.59 & 23.43 & 1.24 & 21.77 \\
FunASR + Kimi-K3 & 9.80 & 9.80 & 9.80 & 10.27 & 0.00 & 11.65 \\
FunASR + Qwen & 5.95 & 5.75 & 5.75 & 5.95 & 0.00 & 5.75 \\
VibeVoice + Kimi-K3 & 11.09 & 11.09 & 11.09 & 12.14 & 3.99 & 11.09 \\
VibeVoice + Qwen & 5.73 & 5.73 & 5.73 & 5.73 & 0.21 & 5.73 \\
\bottomrule
\end{tabular}
\caption{DCIST (\%) under frozen-prediction diagnostics. The first three thresholds divide overlap by predicted duration. Bidirectional coverage additionally requires 50\% of the gold window at the 80\% threshold. ID permutation is globally consistent per recording. Changes are tested separately, retaining semantic and extra-node judgments; 80\% is the official score.}\label{tab:protocol_sensitivity}
\end{table*}

%% file: results_time_windows.tex
\begin{table*}[!t]
\centering
\fontsize{9.5}{11.5}\selectfont
\begin{tabular}{@{}lrrrrr@{}}
\toprule
System & Valid windows & At most 1 s & Median (s) & Matched nodes & Median ratio (\%) \\
\midrule
FunASR + Astra & 732 & 3 & 6.73 & 118 & 25.36 \\
FunASR + Kimi-K3 & 483 & 2 & 5.10 & 46 & 15.54 \\
FunASR + Qwen & 476 & 1 & 5.06 & 23 & 21.54 \\
VibeVoice + Kimi-K3 & 493 & 0 & 10.20 & 50 & 48.56 \\
VibeVoice + Qwen & 489 & 1 & 10.04 & 23 & 29.51 \\
\bottomrule
\end{tabular}
\caption{Predicted evidence-window distribution. The duration ratio is predicted/gold window length (\%) for required nodes in fully matched trajectories. Valid windows have non-negative starts and positive duration.}\label{tab:time_windows}
\end{table*}

%% file: results_noise_localization.tex
\begin{table*}[!t]
\centering
\fontsize{9.5}{11.5}\selectfont
\begin{tabular}{@{}lrrrrrrr@{}}
\toprule
Frontend & Audio & Tracks & Retained & Lost & Gained & Neither & Lost with absent anchor \\
\midrule
FunASR & 26 & 308 & 13 & 26 & 13 & 256 & 26/26 \\
VibeVoice & 22 & 255 & 12 & 16 & 7 & 220 & 13/16 \\
\bottomrule
\end{tabular}
\caption{Paired Kimi-K3 trajectory outcomes for the same-route baseline and B2. Lost means baseline matched and B2 did not; gained is the converse. The final column counts lost trajectories with at least one linked normalized anchor form present before but absent after noise, not proven single-word causes.}\label{tab:noise_localization}
\end{table*}

%% file: astra_script_errors.tex
\begingroup
\expandafter\def\csname @captype\endcsname{table}
\centering
\small
\setlength{\tabcolsep}{5pt}
\renewcommand{\arraystretch}{1.12}
\begin{tabularx}{\textwidth}{@{}>{\raggedright\arraybackslash}p{.235\textwidth}>{\raggedright\arraybackslash}X>{\centering\arraybackslash}p{.155\textwidth}>{\centering\arraybackslash}p{.155\textwidth}@{}}
\toprule
Error type & Criterion & With speaker labels & Without speaker labels \\
\midrule
Action error & \texttt{start / advance / hold / settle} differs from GT & 39 (22.29\%) & 45 (21.63\%) \\
Content error & A number, term, state, or other content in \texttt{value} differs from GT & 94 (53.71\%) & 100 (48.08\%) \\
Speaker error & \texttt{speaker\_id} differs from GT & 38 (21.71\%) & 72 (34.62\%) \\
Scope error & \texttt{scope\_id} differs from GT & 70 (40.00\%) & 92 (44.23\%) \\
Settlement-result error & \texttt{result} is missing, extra, or has an incorrect value & 18 (10.29\%) & 28 (13.46\%) \\
Missing node & A required GT event node is not recovered & 33 (18.86\%) & 42 (20.19\%) \\
Erroneous extra node & An extra output node was judged incorrect in the existing adjudication & 32 (18.29\%) & 43 (20.67\%) \\
Cycle-segmentation error & The number of decision cycles differs from GT & 1 (0.57\%) & 0 (0.00\%) \\
Unaligned field / missing entire trajectory & No corresponding predicted trajectory is found using the current fields and existing mappings & 12 (6.86\%) & 10 (4.81\%) \\
\bottomrule
\end{tabularx}
\caption{Astra's trajectory error distributions under the two reference-script conditions. Entries are the number (percentage) of trajectories involving each error, among 175 unmatched trajectories with speaker labels and 208 without speaker labels.}\label{tab:astra_script_errors}
\vspace{3pt}
\begin{minipage}{\textwidth}
\footnotesize
\textit{Notes.} Each input condition contains 26 scripts and 308 GT trajectories. GT is the reference, and the analysis retains monetary equivalence, existing synonym mappings, and extra-node exemption rules; exempt nodes are not counted as errors. One trajectory may involve multiple error types, so percentages can sum to more than 100\%. Classification first preserves correctly matched nodes, then performs ordered alignment of the unmatched nodes between them. This table covers the \textbf{originally unmatched trajectories}, not the residual trajectories after hypothetically correcting five attribute categories.
\end{minipage}
\endgroup